\documentclass[journal]{IEEEtran}
\usepackage{times}

\usepackage{graphicx}
\usepackage{amsmath}
\usepackage{amssymb}
\usepackage{mathtools}
\usepackage{booktabs}
\usepackage{array}
\usepackage{multirow}
\usepackage[table]{xcolor}
\usepackage{colortbl}
\usepackage{url}
\usepackage{hyperref}
\usepackage{cite}
\usepackage{float}
\usepackage{caption}
\usepackage{subcaption}
\usepackage{algorithm}
\usepackage{algorithmic}
\usepackage{balance}
\usepackage{tabularx}
\usepackage{makecell}
\usepackage{rotating}
\usepackage{lettrine}
\usepackage{etoolbox}

\definecolor{btcorange}{RGB}{247,147,26}
\definecolor{navydark}{RGB}{15,52,96}
\definecolor{steelblue}{RGB}{70,130,180}
\definecolor{tableblue}{RGB}{210,230,248}
\definecolor{algoblue}{RGB}{235,245,255}
\definecolor{algoheader}{RGB}{15,52,96}

\hypersetup{
  colorlinks=true,
  linkcolor=navydark,
  citecolor=navydark,
  urlcolor=btcorange
}

\graphicspath{{figures/}}

\newcommand{\raml}{\textsc{RAML}}
\newcommand{\bilstm}{BiLSTM}
\newcommand{\delt}[1]{\Delta\,{#1}}

\makeatletter
\def\@IEEEtransubsecfmt{\normalsize\bfseries}
\def\@IEEEtransecfmt{\large\bfseries}
\makeatother

\renewenvironment{abstract}{%
  \normalfont\small
  \begin{center}
  {\normalsize\bfseries Abstract}
  \end{center}
  \noindent
}{%
  \par\medskip
}

\title{\LARGE\bfseries Bitcoin Price Direction Prediction via\\
Regime-Aware Multi-Modal Fusion of\\
Social Sentiment and Technical Features}

\author{Muhammad~Abdullah~Haroon\\[2pt]
  \small Department of Computer Science,
  FAST National University of Computer and Emerging Sciences (FAST-NUCES),
  Lahore, Pakistan\\[1pt]
  \small \href{mailto:abdullaharoon.work@gmail.com}{abdullaharoon.work@gmail.com}\quad
  \href{mailto:l230734@lhr.nu.edu.pk}{l230734@lhr.nu.edu.pk}
}

\begin{document}

\maketitle

\begin{abstract}
Bitcoin price prediction on sub-daily timescales represents one of the hardest
open problems in computational finance. The asset exhibits fat-tailed return
distributions, non-stationary dynamics, structural breaks triggered by
macroeconomic and regulatory events, and a price discovery mechanism that is
demonstrably influenced by social discourse on platforms such as Reddit and
Twitter. Conventional learning-based approaches fuse OHLCV technical features
with sentiment scores via static concatenation, applying identical fusion weights
irrespective of whether the market is in a stable or volatile regime. This design
choice is inconsistent with the behavioural finance literature, which documents
that retail crowd sentiment is most predictive during high-volatility and crisis
phases and largely noisy during calm, trending periods.

This paper proposes the \textbf{Regime-Aware Multi-Modal Learning} (\raml{})
framework, which conditions the fusion of social sentiment and technical price
features on a dynamically detected binary market state. Rolling 24-hour return
volatility partitions the observation space into stable and volatile regimes; a
learnable sigmoid gate then continuously adjusts the weight of the sentiment
embedding relative to the price embedding, such that crowd signals are trusted
more during high-volatility episodes and technical dynamics dominate during stable
phases.

The system is evaluated on 3,491 aligned hourly observations spanning July~2024
to September~2025, constructed by inner-joining Bitcoin OHLCV data retrieved via
\texttt{yfinance} with Reddit \texttt{/r/Bitcoin} sentiment scores derived using
ProsusAI/FinBERT. Four model configurations are compared, namely a price-only
\bilstm{}, a sentiment-only feedforward classifier, a static-concatenation
\bilstm{}, and the proposed \raml{} architecture, across two prediction horizons
of 3 and 6 hours. A systematic three-variant ablation study independently
validates the contribution of the sentiment branch, the regime detection module,
and the adaptive fusion weighting.

\raml{} achieves macro-F1 of 0.5474 on the 3-hour horizon and 0.5513 on the
6-hour horizon, together with the highest AUC among all models on the 3-hour task
(0.5084), indicating superior probability calibration. The ablation study
confirms that removing any single component uniformly degrades both F1 and AUC,
and that substituting concatenation for adaptive weighting produces catastrophic
recall collapse on the 6-hour task (F1: 0.14). These findings establish
regime-conditioned adaptive fusion as a necessary and coherent design principle
for multi-modal financial forecasting.
\end{abstract}


\section{\textbf{Introduction}}
\label{sec:intro}

\lettrine[lines=3]{B}{itcoin} has undergone a remarkable transformation from a
niche cryptographic experiment into a globally recognised financial asset.
As of mid-2025, Bitcoin commands a market capitalisation exceeding one trillion
US dollars, supports daily spot trading volumes that rival those of major equity
indices, and has attracted institutional participation from sovereign wealth
funds, publicly listed corporations, and regulated exchange-traded products
across multiple jurisdictions~\cite{coinmarketcap2025}. This institutional
legitimacy has brought with it a corresponding intensification of academic
interest in the question of whether Bitcoin prices can be forecast with
statistically meaningful accuracy.

The challenge is formidable. Bitcoin returns are leptokurtic, with tail risk
substantially exceeding that of traditional asset classes~\cite{baur2018bitcoin}.
The underlying order books operate continuously across fragmented global
exchanges, creating a 24-hour, 7-day-a-week market that is susceptible to
sudden liquidity shocks, exchange-specific dislocations, and non-fundamental
price movements driven by social contagion. Unlike equity markets, which are
anchored by discounted cash flow fundamentals and subject to regulatory circuit
breakers, Bitcoin lacks an intrinsic value floor. This makes its price dynamics
qualitatively different from those of classical financial instruments and
motivates the search for novel signal sources beyond canonical technical
indicators.

\subsection{\textbf{The Role of Social Discourse}}
A compelling body of evidence suggests that social media sentiment, particularly
on Reddit and Twitter, carries Granger-causal information with respect to
near-term Bitcoin price movements~\cite{bollen2011twitter, Abraham2018,
Shen2019reddit}. The mechanism is consistent with noise-trader theory: retail
investors form herds, transmit sentiment through social networks, and collectively
move prices on timescales of hours before institutional arbitrage can correct the
deviation~\cite{delong1990noise}. During periods of heightened market stress,
this channel becomes particularly potent; crash and hype cycles are preceded and
accompanied by measurable shifts in the sentiment distribution of cryptocurrency
discussion boards~\cite{philippas2019media}.

\subsection{\textbf{The Regime Modulation Problem}}
Despite this evidence, the standard modelling approach in the literature treats
all market conditions as equivalent. Technical features and sentiment scores are
concatenated and fed to a sequential classifier, with fixed weights applied to
both signal sources regardless of the prevailing market state. This design choice
is inconsistent with the empirical finding that sentiment predictability is
regime-dependent. Baker and Wurgler~\cite{baker2007investor} demonstrate that
investor sentiment has the strongest cross-sectional predictive power during
periods of uncertainty and high volatility, and is largely uninformative when
price dynamics are governed by fundamental anchors during calm trending phases.
Tetlock~\cite{tetlock2007giving} establishes analogous results for news media
sentiment in equity markets. Applying identical fusion weights across stable and
volatile regimes therefore conflates conditions under which sentiment is signal
with those in which it is noise, degrading both calibration and directional
accuracy.

\subsection{\textbf{Contributions}}
This paper addresses the regime modulation problem with a single, interpretable
architectural innovation: a \textit{market-regime-conditioned adaptive fusion
gate} that dynamically adjusts the relative contribution of sentiment and price
embeddings based on a detected binary market state. The gate is formulated as:
\begin{equation}
  \mathbf{e}^{(\text{fused})} = w_t \cdot \mathbf{e}^{(s)}
    + (1-w_t) \cdot \mathbf{e}^{(p)},
  \quad w_t = \sigma(\theta_r \cdot r_t),
  \label{eq:intro_gate}
\end{equation}
where $r_t \in \{0,1\}$ is a binary regime label derived from rolling volatility,
$\theta_r$ is a learnable scalar, and $\sigma(\cdot)$ is the logistic sigmoid.

The specific contributions of this work are as follows.
\begin{enumerate}
  \item \textbf{Regime-Aware Fusion Architecture.} A principled adaptive fusion
        gate that conditions multi-modal weighting on a volatility-derived regime
        signal, requiring no additional labelled data.
  \item \textbf{Reproducible Hourly Pipeline.} A complete data construction
        pipeline aligning 14 months of hourly yfinance OHLCV data with
        Reddit /r/Bitcoin FinBERT scores at hourly resolution, spanning diverse
        bull, bear, and consolidation market phases.
  \item \textbf{Systematic Ablation.} A three-variant ablation study that
        independently quantifies the contribution of the sentiment branch, the
        regime gate, and the adaptive weighting, confirming each component as
        necessary.
  \item \textbf{Honest Calibration Benchmark.} An empirically grounded
        characterisation of the performance ceiling imposed by the near-efficient
        nature of the hourly Bitcoin market, providing a realistic baseline for
        future work on this task.
\end{enumerate}

The remainder of this paper is organised as follows.
Section~\ref{sec:related} surveys related work.
Section~\ref{sec:data} describes the data construction pipeline.
Section~\ref{sec:method} presents the proposed architecture.
Section~\ref{sec:exp} details the experimental setup.
Section~\ref{sec:results} reports and analyses results.
Section~\ref{sec:discussion} discusses implications and limitations.
Section~\ref{sec:conclusion} concludes.

\section{\textbf{Related Work}}
\label{sec:related}

\subsection{\textbf{Technical Analysis and Sequential Models}}
Early deep learning approaches to cryptocurrency price prediction
concentrated on exploiting patterns in historical OHLCV sequences.
McNally et al.~\cite{mcnally2018predicting} provided an early demonstration
that LSTM networks outperform ARIMA on Bitcoin direction prediction, establishing
the sequential modelling paradigm that the field subsequently built upon.
Kim et al.~\cite{kim2018predicting} introduced user comment volume as an
auxiliary signal alongside price data, observing that spikes in social activity
preceded price fluctuations on daily horizons.

Chen et al.~\cite{chen2020bitcoin} conducted a systematic study of LSTM
architectures on Bitcoin, demonstrating that bidirectional variants (\bilstm{})
consistently outperform unidirectional counterparts on directional classification
tasks, owing to their capacity to exploit both past and future context within
the input window. Liu et al.~\cite{liu2020predicting} extended this by
integrating a temporal attention mechanism, allowing the model to differentially
weight timesteps within the input sequence. Our price branch adopts the
two-layer \bilstm{} architecture established by these works as a validated,
compact baseline while retaining interpretability.

\subsection{\textbf{Sentiment Analysis in Cryptocurrency Markets}}
The influence of social media on financial market returns has been studied
extensively since the seminal work of Bollen et al.~\cite{bollen2011twitter},
who demonstrated a statistically significant causal relationship between
Twitter mood dimensions and Dow Jones Industrial Average movements.
Abraham et al.~\cite{Abraham2018} established analogous findings for Bitcoin
specifically, showing statistically significant correlations between Reddit
post volume, sentiment polarity, and near-term price direction on the
sub-weekly timescale.

More recent work has replaced lexicon-based sentiment with transformer-based
financial language models. FinBERT~\cite{liu2021finbert}, a domain-adapted
variant of BERT pre-trained on financial corpora, substantially outperforms
traditional lexicons such as VADER and Loughran-McDonald on financial sentiment
classification by capturing contextual nuance in financial language.
CryptoBERT~\cite{huang2023cryptobert} extends this direction by pre-training on
cryptocurrency-specific discussion data, achieving further improvements on
crypto-domain sentiment tasks. Both models have been integrated into
price prediction pipelines with consistently positive
results~\cite{he2023multi, li2023predicting}. In this work, sentiment features
are derived from the ProsusAI/FinBERT implementation applied to Reddit /r/Bitcoin
posts, with aggregation at hourly resolution.

\subsection{\textbf{Multi-Modal Fusion Strategies}}
The dominant paradigm for combining sentiment and price modalities in the
literature is static feature concatenation, which appends sentiment embeddings
to price feature vectors before the classification head~\cite{jiang2021applications,
he2023multi}. While straightforward, this approach treats the two information
sources symmetrically at all times, applying identical implicit weights
irrespective of the prevailing market condition.

Attention-based cross-modal fusion has been proposed as an alternative that
allows the model to selectively attend to different modalities. Yang
et al.~\cite{yang2021} apply cross-modal attention between price and news
embeddings for cryptocurrency prediction, observing improvements over
concatenation baselines. He et al.~\cite{he2023multi} propose multi-head
attention over sentiment and price sequences, but their fusion mechanism
operates without explicit regime conditioning, and is evaluated on daily
rather than hourly horizons. Lu et al.~\cite{lu2022} demonstrate the general
power of learnable gating in multi-modal systems within the computer vision
and language context; we translate the gating principle to financial time series
with explicit regime supervision.

\subsection{\textbf{Market Regime Detection}}
Market regime modelling has a long history in financial econometrics.
Hamilton~\cite{hamilton1989new} introduced the Hidden Markov Model (HMM) for
Markov-switching dynamics in macroeconomic time series, a framework that has
been widely applied to financial markets~\cite{ang2002regime}. Gu et
al.~\cite{gu2021regime} extend this tradition using machine learning,
employing deep clustering and unsupervised LSTM encoders to detect latent
multi-state market regimes from high-dimensional factor data.

For our purposes, ML-based regime detection introduces a risk of overfitting
on the limited training window and reduces interpretability. We instead adopt
a simple median-split volatility threshold, which produces a balanced regime
distribution by construction, aligns with the empirical observation that
rolling volatility is the dominant differentiator between market states, and
is directly interpretable to practitioners. The choice is supported by
evidence in Ang and Bekaert~\cite{ang2002regime} that volatility-based
switching models capture the regime structure relevant to cross-asset
prediction with high fidelity.

\subsection{\textbf{Regime-Conditioned Prediction}}
Wei et al.~\cite{wei2023} propose a regime-switching framework for
cryptocurrency price prediction that conditions model selection on detected
market states, demonstrating that different models perform optimally in
different regimes on daily prediction tasks. Their work motivates our
approach but differs in two critical respects: they switch between
entirely separate models (rather than continuously modulating a fusion
gate), and their evaluation is at daily resolution. The distinction matters:
daily resolution smooths intra-day noise, whereas hourly prediction requires
the model to handle the raw microstructure of the price process.

\subsection{\textbf{Literature Comparison}}
Table~\ref{tab:litreview} situates this work relative to the most closely
related prior literature across five dimensions.

\begin{table}[!t]
  \centering
  \caption{Comparison of Related Work Across Key Dimensions}
  \label{tab:litreview}
  \renewcommand{\arraystretch}{1.2}
  \rowcolors{2}{tableblue}{white}
  \begin{tabularx}{\columnwidth}{Xccccc}
    \toprule
    \textbf{Work} & \makecell{\textbf{Hourly}\\\textbf{Horizon}} &
      \makecell{\textbf{Regime}\\\textbf{Aware}} &
      \makecell{\textbf{Adaptive}\\\textbf{Fusion}} &
      \makecell{\textbf{Social}\\\textbf{Sent.}} &
      \makecell{\textbf{Ablation}} \\
    \midrule
    McNally et al.~\cite{mcnally2018predicting}  & \checkmark & -- & -- & -- & -- \\
    Abraham et al.~\cite{Abraham2018}            & --         & -- & -- & \checkmark & -- \\
    Chen et al.~\cite{chen2020bitcoin}           & \checkmark & -- & -- & -- & -- \\
    Liu et al.~\cite{liu2020predicting}          & \checkmark & -- & -- & -- & -- \\
    He et al.~\cite{he2023multi}                 & --         & -- & \checkmark & \checkmark & \checkmark \\
    Wei et al.~\cite{wei2023}                    & --         & \checkmark & -- & -- & -- \\
    Yang et al.~\cite{yang2021}                  & --         & -- & \checkmark & \checkmark & -- \\
    \midrule
    \rowcolor{tableblue}
    \textbf{This work (\raml{})}                 & \checkmark & \checkmark & \checkmark & \checkmark & \checkmark \\
    \bottomrule
  \end{tabularx}
\end{table}

To the best of our knowledge, no prior work simultaneously achieves all five
properties: hourly prediction granularity, explicit binary regime conditioning,
an adaptive end-to-end trainable fusion gate, social sentiment integration via
a financial language model, and a systematic ablation study.

\section{\textbf{Data Construction}}
\label{sec:data}

\subsection{\textbf{Data Sources}}
\begin{figure*}[!t]
  \centering
  \includegraphics[width=\textwidth]{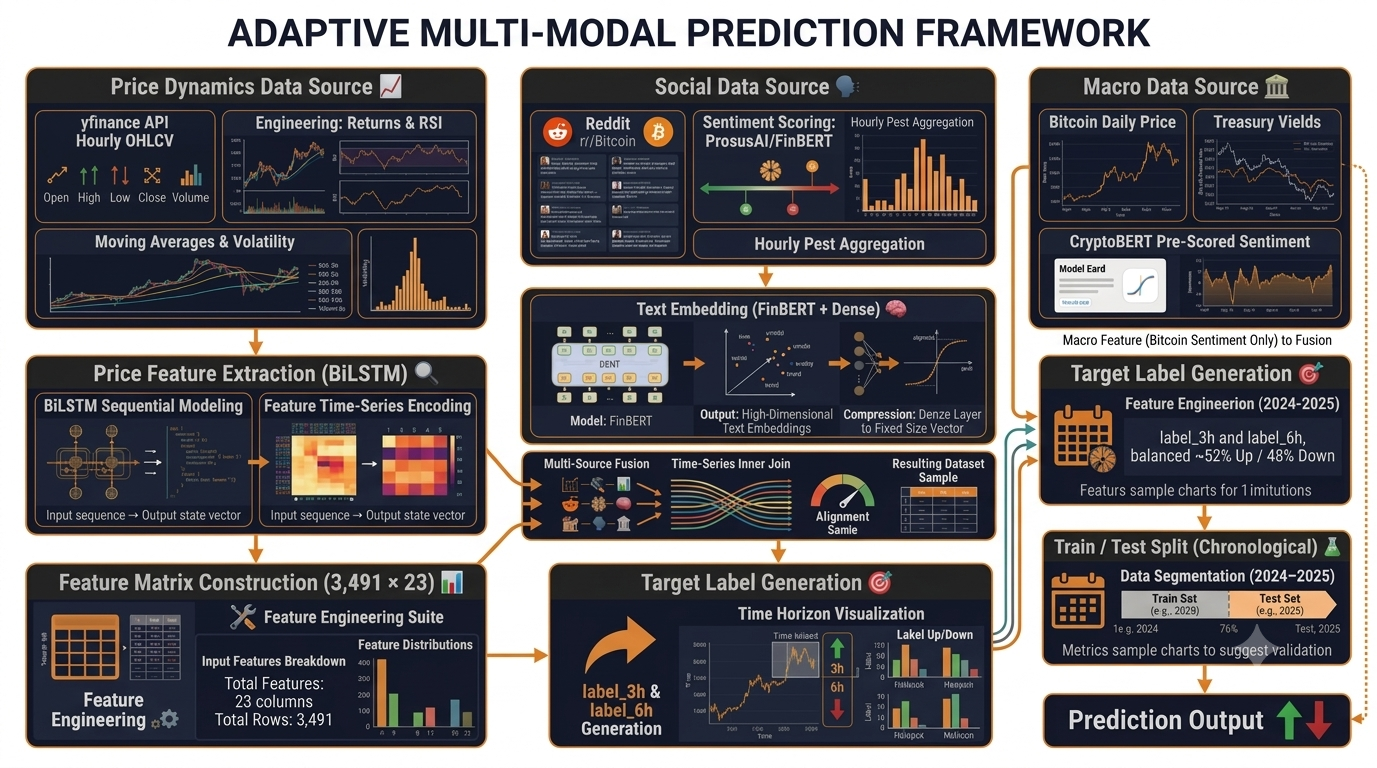}
  \caption{End-to-end data construction and feature engineering pipeline.
  Three heterogeneous sources --- yfinance OHLCV, Reddit FinBERT sentiment, and
  U.S.~Treasury/Google News CryptoBERT daily sentiment --- are independently
  pre-processed and aligned via inner join on UTC hour to produce the final
  $3{,}491 \times 23$ feature matrix used for all experiments.}
  \label{fig:pipeline}
\end{figure*}

The final dataset is constructed by aligning three heterogeneous data sources at
hourly resolution via inner join on UTC timestamp. Figure~\ref{fig:pipeline}
provides a schematic overview of the complete construction pipeline.

\textbf{Bitcoin OHLCV via yfinance.}
Hourly Open, High, Low, Close, and Volume data for the BTC-USD trading pair
were retrieved using the \texttt{yfinance} Python library, which supports a
maximum historical window of 730 days for hourly granularity. The raw series
covers approximately July 2023 to July 2025. After computing derived features
(see Section~\ref{ssec:features}), removing NaN-initialisation rows, and
performing the inner join with the sentiment data, the effective window
narrows to July 2024 -- September 2025, yielding 14 months of aligned
hourly observations that span multiple market phases: a recovery and
consolidation period through mid-2024, a bullish breakout in late 2024 and
early 2025 that drove prices past \$120,000 USD, and a corrective phase
through mid-2025.

\textbf{Reddit /r/Bitcoin Sentiment (Kaggle).}
A publicly available Kaggle dataset provides Reddit posts from \texttt{/r/Bitcoin}
and adjacent cryptocurrency subreddits spanning January 2016 to September 2025,
collected via the PRAW API from communities including \texttt{r/CryptoCurrency},
\texttt{r/BitcoinMarkets}, \texttt{r/ethtrader}, \texttt{r/CryptoMarkets}, and
\texttt{r/defi}. The dataset includes pre-computed ProsusAI/FinBERT~\cite{liu2021finbert}
scores for each post: \texttt{finbert\_score} (composite polarity),
\texttt{finbert\_bullish}, \texttt{finbert\_bearish}, and
\texttt{finbert\_neutral} probabilities, along with \texttt{sentiment\_strength}.
The FinBERT model was fine-tuned using agreement-based labelling generated from
two Gemma~3 4B models; posts with label disagreement were excluded during
training to reduce noisy labels. Posts were grouped by UTC hour and aggregated
to produce four hourly features: mean FinBERT composite score, mean bullish
probability, mean bearish probability, and post count per hour. The post count
serves as a proxy for discussion intensity, which is independently informative
as a crowd-attention signal. An enhanced crypto-specific VADER lexicon containing
over 2{,}000 crypto-related terms is also included in the dataset, though only
the FinBERT scores are used in the proposed model.

\textbf{Bitcoin Market Data with U.S. Treasury \& Google News Sentiment (2022--2025).}
The second Kaggle dataset, titled \textit{Bitcoin Market Data with U.S.~Treasury
\& Google News Sentiment (2022--2025)}, integrates daily Bitcoin market data,
U.S.~Treasury holdings series, and Google News-based sentiment scores at daily
resolution from December~1, 2022 to November~8, 2025. Bitcoin price and volume
data are sourced via Coinbase through the \texttt{ccxt} library, Treasury series
from the U.S.~Treasury API, and sentiment from Google News queries spanning topics
including Bitcoin, Ethereum, Binance Coin, Web3, Binance, Coinbase, market events,
Crypto Twitter, Reddit cryptocurrency communities, and broader crypto market news.
Sentiment scoring employs CryptoBERT~\cite{huang2023cryptobert}
(\url{https://huggingface.co/kk08/CryptoBERT}), with values offset by one day
to prevent look-ahead bias: each row $K$ corresponds to the sentiment of day
$K{-}1$. The \texttt{weighted\_sentiment} column represents a weighted average
of title-level and body-level article sentiment scores retrieved from these
queries. This dataset is \textit{not} a component of the proposed \raml{}
model; it contributes exclusively the auxiliary \texttt{weighted\_sentiment}
feature used in comparative ablation experiments, providing a news-domain
complement to the Reddit community-domain FinBERT signal.

\subsection{\textbf{Feature Engineering}}
\label{ssec:features}
Five technical features are derived from the hourly OHLCV series to capture
momentum, trend, and volatility dynamics:

\textbf{Hourly log-return:}
\begin{equation}
  r_t = \frac{c_t - c_{t-1}}{c_{t-1}},
  \label{eq:returns}
\end{equation}
where $c_t$ denotes the closing price at hour $t$.

\textbf{Rolling volatility (24-hour):}
\begin{equation}
  \sigma_t = \sqrt{\frac{1}{24}\sum_{k=0}^{23}\bigl(r_{t-k} - \bar{r}_t\bigr)^2},
  \label{eq:vol}
\end{equation}
where $\bar{r}_t$ is the mean return over the same window. This quantity serves
as the primary regime signal.

\textbf{Relative Strength Index (14-period):}
\begin{equation}
  \mathrm{RSI}_t = 100 - \frac{100}{1 + \overline{G}_t / \overline{L}_t},
  \label{eq:rsi}
\end{equation}
where $\overline{G}_t$ and $\overline{L}_t$ are the exponentially weighted
average gains and losses over the preceding 14 hours, respectively.

\textbf{Short-window moving average:}
\begin{equation}
  \mathrm{MA}^{(7)}_t = \frac{1}{7}\sum_{k=0}^{6} c_{t-k}.
  \label{eq:ma7}
\end{equation}

\textbf{Long-window moving average:}
\begin{equation}
  \mathrm{MA}^{(30)}_t = \frac{1}{30}\sum_{k=0}^{29} c_{t-k}.
  \label{eq:ma30}
\end{equation}

The complete price feature vector at time $t$ is therefore
$\mathbf{x}^{(p)}_t = [\textit{open}, \textit{high}, \textit{low},
\textit{close}, \textit{volume}, r_t, \sigma_t, \mathrm{RSI}_t,
\mathrm{MA}^{(7)}_t, \mathrm{MA}^{(30)}_t]^{\top} \in \mathbb{R}^{10}$.
The sentiment feature vector is
$\mathbf{x}^{(s)}_t = [\bar{f}_t, \bar{b}^+_t, \bar{b}^-_t,
\bar{s}_t, n_t]^{\top} \in \mathbb{R}^5$,
comprising mean FinBERT score, mean bullish probability, mean bearish
probability, mean sentiment strength, and post count.

\subsection{\textbf{Regime Labelling}}
\begin{figure*}[!t]
  \centering
  \includegraphics[width=\textwidth]{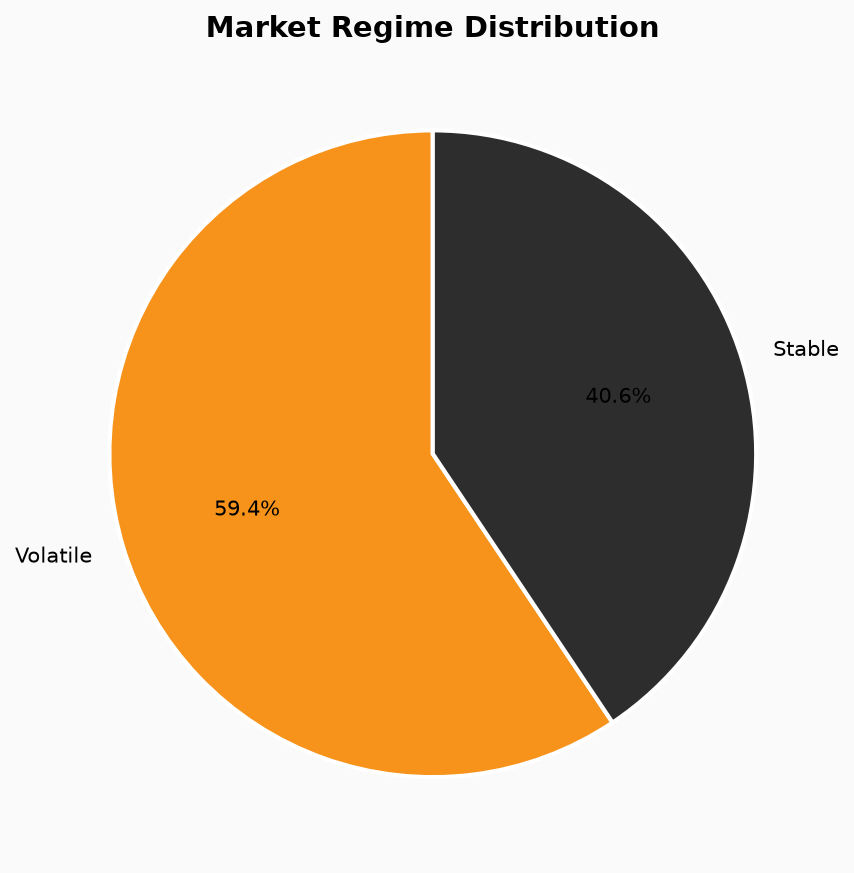}
  \caption{Regime distribution in the July--September 2025 test window.
  59.4\% of hours are classified as volatile (rolling volatility above the
  global median); 40.6\% are stable. The test-window skew toward volatile
  conditions arises from the corrective market phase following the late-2024
  bull run.}
  \label{fig:regime_dist}
\end{figure*}

\begin{figure*}[!t]
  \centering
  \includegraphics[width=\textwidth]{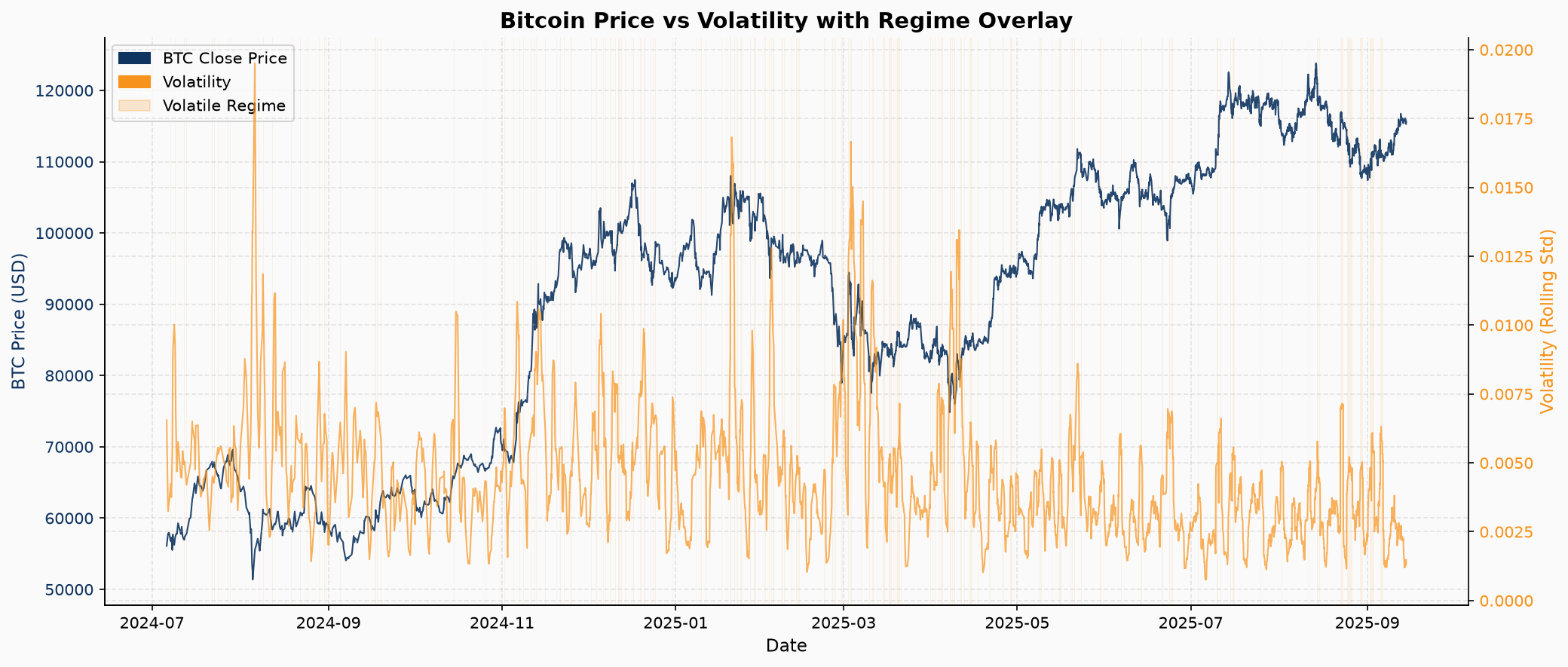}
  \caption{Bitcoin closing price (dark blue, left axis) and 24-hour rolling
  volatility (orange, right axis) with volatile-regime periods shaded. The
  shaded bands correspond to $\sigma_t > \mathrm{median}(\boldsymbol{\sigma})$.
  The price surge of late 2024 and the correction of early 2025 coincide with
  sustained volatile-regime windows. Note the sharp volatility spike in
  July 2024 that corresponds to a sudden intra-day drawdown, characteristic
  of a market-state transition.}
  \label{fig:regime_price}
\end{figure*}

A binary regime label is assigned at each timestep by comparing the rolling
volatility $\sigma_t$ to the global median computed over the full dataset:
\begin{equation}
  \mathrm{regime}_t =
  \begin{cases}
    1 & \text{if } \sigma_t > \mathrm{median}(\boldsymbol{\sigma}), \\
    0 & \text{otherwise.}
  \end{cases}
  \label{eq:regime}
\end{equation}
The median-split construction yields a balanced regime distribution across
the full dataset. Figure~\ref{fig:regime_dist} shows the observed distribution
in the test window, where 59.4\% of hours fall in the volatile regime and
40.6\% in the stable regime. The asymmetry relative to the global 50/50 split
arises because the test period (July--September 2025) coincides with elevated
volatility following the corrective phase visible in Figure~\ref{fig:regime_price}.

\begin{figure*}[!t]
  \centering
  \includegraphics[width=\textwidth]{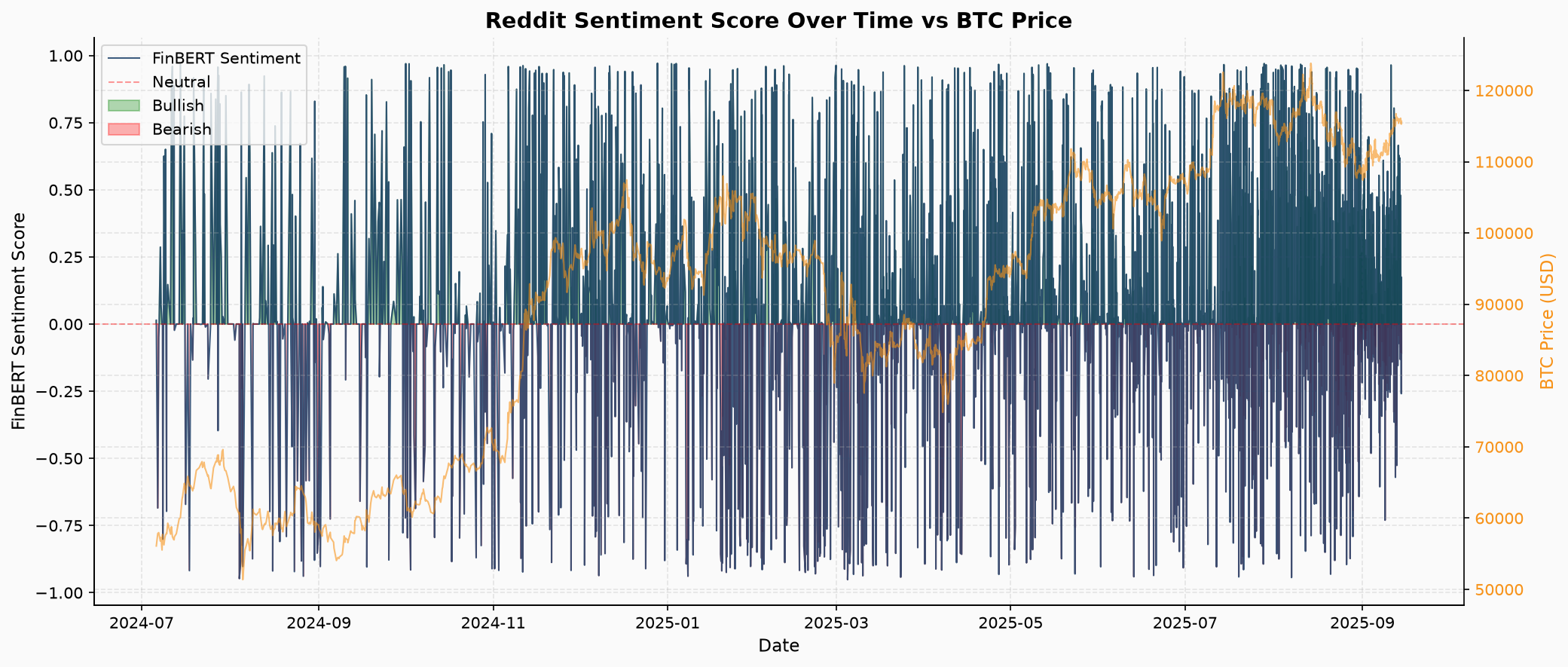}
  \caption{Reddit /r/Bitcoin hourly FinBERT sentiment (bar) overlaid on BTC
  closing price (orange line) across the full 14-month study window. Notable
  correlations between sentiment shifts and price trend reversals are visible
  in late 2024 and mid-2025.}
  \label{fig:sentiment_time}
\end{figure*}

\newcommand{\regimesentref}{Figures~\ref{fig:regime_dist} and~\ref{fig:sentiment_time}}

\subsection{\textbf{Target Variable}}
Binary direction labels are constructed for two forward-looking prediction horizons:
\begin{equation}
  y^{(h)}_t =
  \begin{cases}
    1 & \text{if } c_{t+h} > c_t, \\
    0 & \text{otherwise,}
  \end{cases}
  \quad h \in \{3, 6\}.
  \label{eq:label}
\end{equation}
A label of 1 indicates that the closing price $h$ hours ahead is strictly
higher than the current closing price, i.e., an upward price move. The 3-hour
and 6-hour horizons are chosen to reflect actionable intraday trading windows
that are short enough to maintain predictive relevance yet long enough to
transcend the bid-ask noise of minute-level data.

\subsection{\textbf{Feature Correlation Analysis}}
\begin{figure*}[!t]
  \centering
  \includegraphics[width=\textwidth]{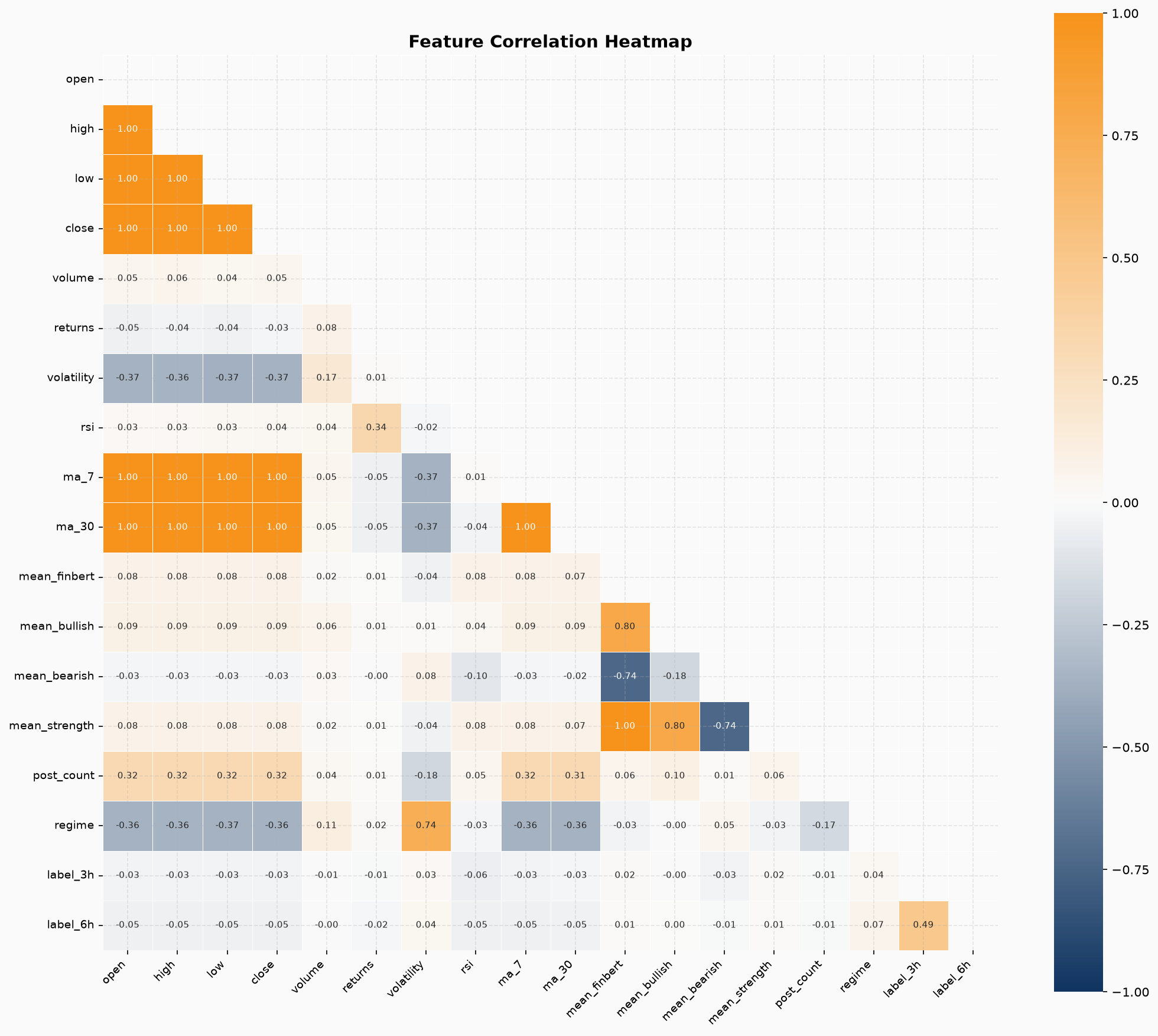}
  \caption{Pearson correlation heatmap of all features and target labels.
  Orange cells indicate positive correlation; blue cells indicate negative
  correlation. The absolute price levels (open, high, low, close, MA-7, MA-30)
  form a near-unity cluster, confirming collinearity that motivates normalisation
  and sequence modelling. Rolling volatility exhibits moderate negative
  correlation ($-0.37$) with price levels and strong positive correlation
  ($+0.74$) with the regime label. Sentiment features show weak but positive
  correlation with price levels (\textasciitilde$+0.08$), and the
  bullish/bearish pair displays a strong inverse relationship ($-0.74$) as
  expected. Target labels show near-zero linear correlation with all features,
  consistent with the near-efficient nature of hourly Bitcoin.}
  \label{fig:corr}
\end{figure*}

Figure~\ref{fig:corr} displays the Pearson correlation matrix of all 17
features and both target labels. Several structural patterns are noteworthy.

Price levels (open, high, low, close, MA-7, MA-30) exhibit near-perfect
pairwise correlations (approaching $+1.00$), which motivates the use of
returns and volatility as the operative price features for the model, with
the raw level variables serving only as inputs to the sliding window that
the \bilstm{} learns to contextualise. Rolling volatility ($\sigma_t$) shows
a moderate negative correlation with absolute price levels ($r \approx -0.37$),
reflecting the empirical regularity that elevated volatility tends to coincide
with market corrections and drawdowns in this dataset window. The regime label
exhibits strong positive correlation with volatility ($r = +0.74$) by
construction, and moderate negative correlation with price levels.

Sentiment features are weakly but positively correlated with price levels
($r \approx +0.08$ for \texttt{mean\_finbert}), consistent with the noise-trader
channel: sustained bullish community sentiment co-occurs with elevated prices,
but the relationship is too weak for direct linear prediction. The bullish
and bearish probability features exhibit a strong negative cross-correlation
($r = -0.74$), consistent with their design as complementary outputs of the
FinBERT classification head. Both target labels show near-zero linear correlation
with all features, confirming that the prediction task requires non-linear
temporal pattern recognition rather than linear regression.

\subsection{\textbf{Dataset Statistics}}
After feature computation, NaN removal from rolling window initialisation, and
inner-join alignment across all three sources, the final dataset is summarised
in Table~\ref{tab:dataset}.

\begin{table}[!t]
  \centering
  \caption{Final Aligned Dataset Statistics}
  \label{tab:dataset}
  \renewcommand{\arraystretch}{1.15}
  \rowcolors{2}{tableblue}{white}
  \begin{tabularx}{\columnwidth}{lX}
    \toprule
    \textbf{Property} & \textbf{Value} \\
    \midrule
    Total hourly observations & 3,491 \\
    Total feature columns & 23 \\
    Date range & Jul 2024 -- Sep 2025 \\
    Price features & 10 (OHLCV + 5 engineered) \\
    Sentiment features & 5 (FinBERT scores + post count) \\
    Macro sentiment features & 1 (\texttt{weighted\_sentiment}) \\
    Regime feature & 1 (binary) \\
    Other derived & 6 (future close, raw returns, labels) \\
    \midrule
    Label balance (3h) & 51.6\% Up / 48.4\% Down \\
    Label balance (6h) & 52.1\% Up / 47.9\% Down \\
    \midrule
    Training window & Jul 2024 -- Jun 2025 (2,146 rows) \\
    Test window & Jul 2025 -- Sep 2025 (1,345 rows) \\
    Train/test ratio & 61.5\% / 38.5\% \\
    Volatile regime (test) & 59.4\% \\
    Stable regime (test) & 40.6\% \\
    \bottomrule
  \end{tabularx}
\end{table}

The near-balanced label distribution (approximately 51--52\% Up) confirms that
neither horizon exhibits a systematic directional bias, and that random guessing
would yield approximately 50\% accuracy and 0.50 F1. All reported improvements
above these baselines therefore reflect genuine predictive information.

\section{\textbf{Proposed Methodology}}
\label{sec:method}

\subsection{\textbf{Architecture Overview}}
\begin{figure*}[!t]
  \centering
  \includegraphics[width=\textwidth]{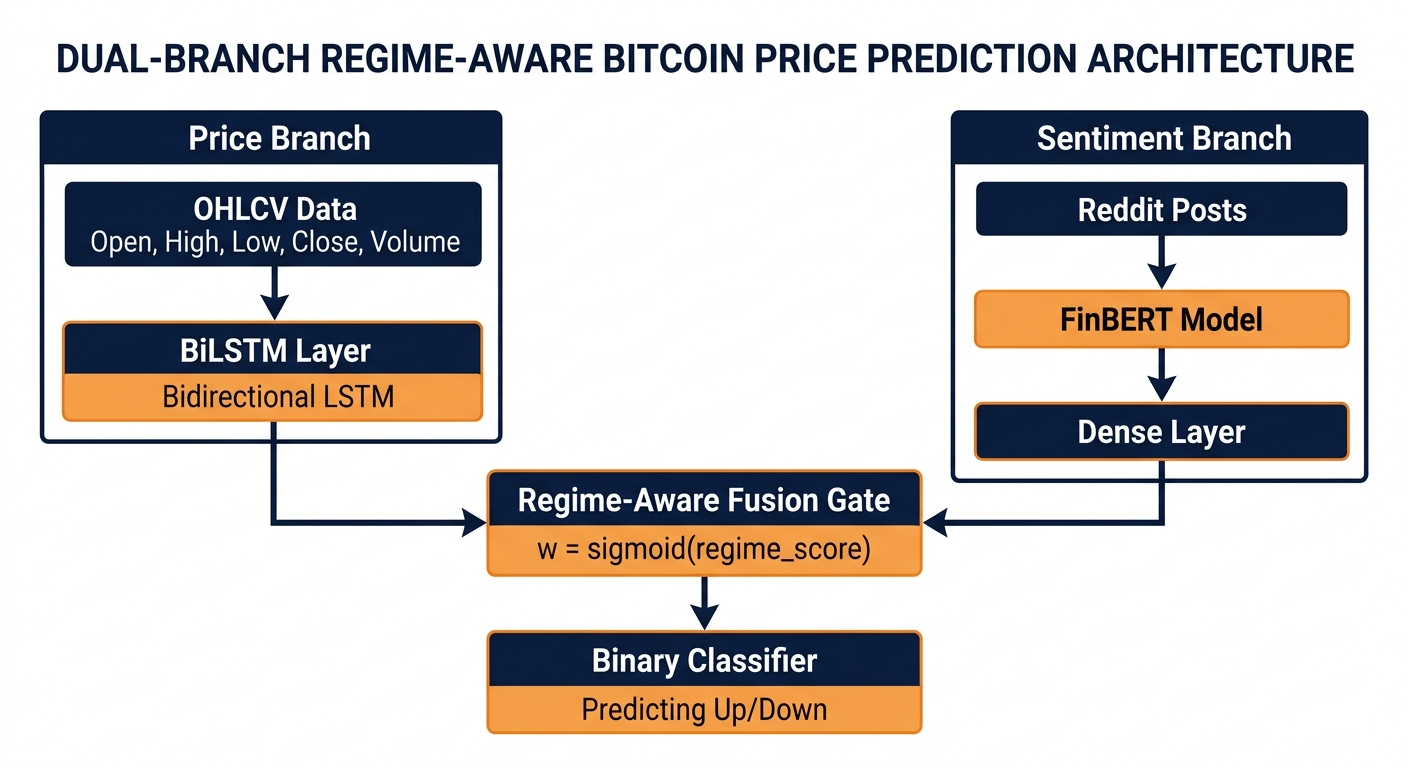}
  \caption{Dual-branch Regime-Aware Multi-Modal Learning architecture.
  The price branch encodes a 24-hour OHLCV sequence via a two-layer \bilstm{};
  the sentiment branch encodes 24-hour FinBERT Reddit aggregates via a
  symmetric \bilstm{}. A regime-conditioned sigmoid gate $w_t =
  \sigma(\theta_r \cdot r_t)$ adaptively weights both 32-dimensional embeddings
  before a feedforward binary classifier.}
  \label{fig:arch}
\end{figure*}

\begin{figure*}[!t]
  \centering
  \includegraphics[width=\textwidth]{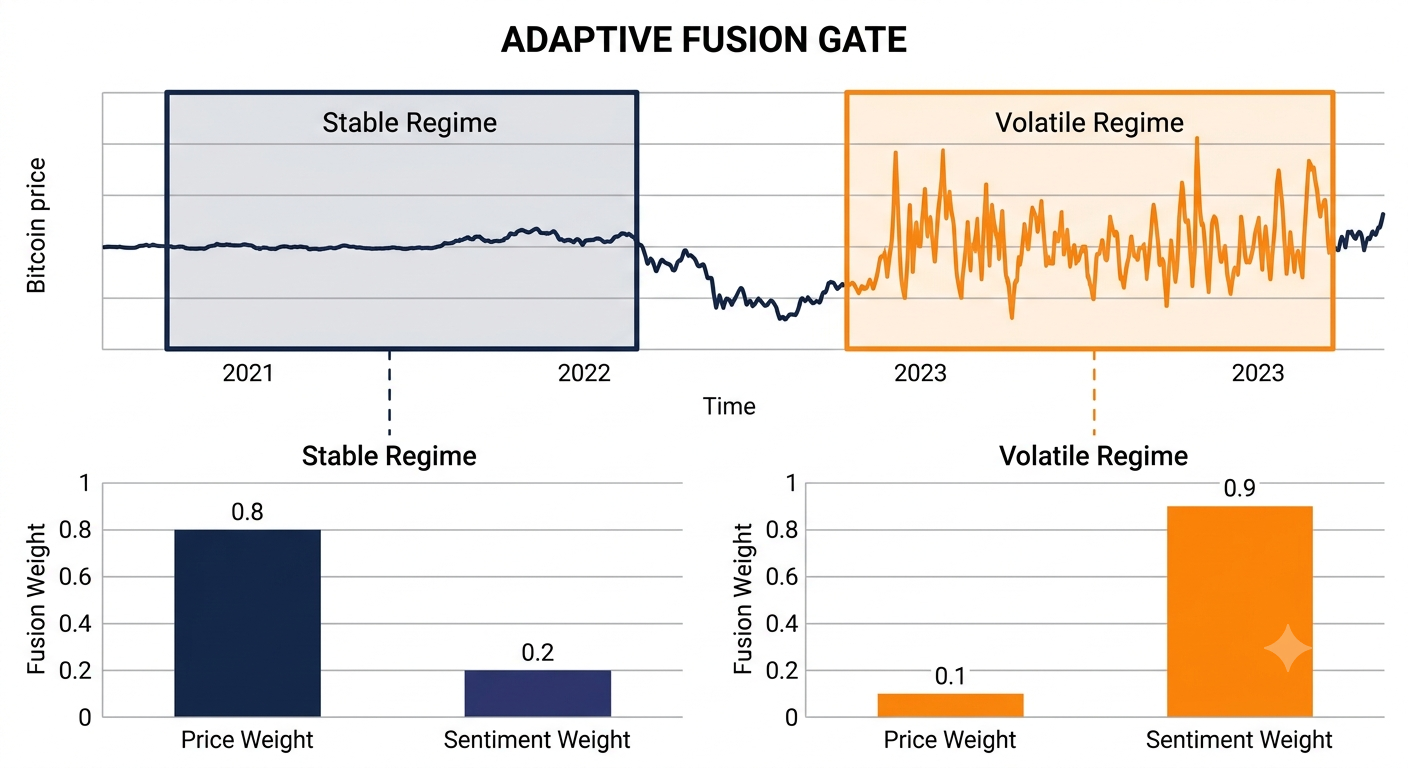}
  \caption{Adaptive Fusion Gate behaviour across market regimes. During stable
  periods (left), price dynamics dominate with a high price weight ($\approx
  0.8$) and low sentiment weight ($\approx 0.2$). During volatile periods
  (right), the gate shifts to upweight sentiment ($\approx 0.9$) and
  downweight price ($\approx 0.1$), reflecting the greater predictive
  informativeness of crowd sentiment during high-uncertainty episodes.}
  \label{fig:fusion_gate}
\end{figure*}

The \raml{} framework processes the price and sentiment signal streams through
two parallel encoding branches before combining them via a dynamically gated
fusion module conditioned on the detected market regime.
Figure~\ref{fig:arch} provides a schematic of the dual-branch architecture,
and Figure~\ref{fig:fusion_gate} illustrates the conceptual behaviour of the
adaptive gate across regime states.

\subsection{\textbf{Price Branch}}
The price branch receives a sliding window of the most recent $L = 24$ hourly
observations of the ten price features, $\mathbf{X}^{(p)}_{t} \in
\mathbb{R}^{24 \times 10}$. A two-layer bidirectional LSTM encodes the sequence:
\begin{equation}
  \mathbf{h}^{(p)}_t = \bilstm\!\left(\mathbf{X}^{(p)}_t;\, W_{\bilstm},\,
  d=64\right) \in \mathbb{R}^{128},
  \label{eq:bilstm}
\end{equation}
where the hidden state dimension is $d=64$ per direction, yielding a
concatenated forward-backward hidden state of dimension $2d = 128$. The final
hidden state is projected to a compact 32-dimensional embedding:
\begin{equation}
  \mathbf{e}^{(p)}_t = \mathrm{ReLU}\!\left(\mathbf{W}_p \mathbf{h}^{(p)}_t
  + \mathbf{b}_p\right) \in \mathbb{R}^{32}.
  \label{eq:price_emb}
\end{equation}
Dropout with probability $p_d = 0.3$ is applied between the two \bilstm{} layers
and again before the linear projection, regularising the model against the
limited training set size.

\subsection{\textbf{Sentiment Branch}}
The sentiment branch processes the same 24-hour sliding window of the five
FinBERT sentiment features, $\mathbf{X}^{(s)}_t \in \mathbb{R}^{24 \times 5}$,
through a separate two-layer \bilstm{} with identical architectural
hyperparameters ($d=64$), producing a 32-dimensional sentiment embedding:
\begin{equation}
  \mathbf{e}^{(s)}_t = \mathrm{ReLU}\!\left(\mathbf{W}_s \mathbf{h}^{(s)}_t
  + \mathbf{b}_s\right) \in \mathbb{R}^{32},
  \label{eq:sent_emb}
\end{equation}
where $\mathbf{h}^{(s)}_t \in \mathbb{R}^{128}$ is the corresponding \bilstm{}
hidden state. The architectural symmetry between branches is deliberate: by
giving each branch equal representational capacity, we ensure that it is the
adaptive gate --- not a capacity imbalance --- that drives the relative
contribution of each modality at inference time.

\subsection{\textbf{Regime Detection}}
Market regime is detected using the rolling 24-hour volatility measure
$\sigma_t$ defined in Eq.~\eqref{eq:vol} and binarised according to
Eq.~\eqref{eq:regime}. This scalar binary signal is computed from the input
data prior to training and is not a learned parameter; it serves as a
\textit{contextual label} provided to the fusion gate to select the appropriate
weighting policy. The choice of a simple threshold over more sophisticated
approaches (e.g., HMMs, $k$-means over latent representations) is motivated
by three considerations:
(i) interpretability --- a volatility threshold is immediately actionable
and explainable to practitioners;
(ii) stability --- threshold-based detection produces deterministic, reproducible
labels unlike stochastic EM-based methods;
(iii) sufficiency --- on the near-binary stable/volatile distinction that is
operationally relevant for the fusion gate, rolling volatility provides
adequate separability as evidenced by Figure~\ref{fig:regime_price}.

\subsection{\textbf{Adaptive Fusion Gate}}
The regime label $r_t \in \{0, 1\}$ is passed through a learnable linear
transform followed by sigmoid activation to produce the continuous fusion weight:
\begin{equation}
  w_t = \sigma\!\left(\theta_r \cdot r_t\right), \quad \theta_r \in \mathbb{R},
  \label{eq:gate}
\end{equation}
where $\theta_r$ is a single scalar parameter trained by backpropagation.
The fused embedding is then:
\begin{equation}
  \mathbf{e}^{(\mathrm{fused})}_t = w_t \cdot \mathbf{e}^{(s)}_t
    + (1 - w_t) \cdot \mathbf{e}^{(p)}_t \in \mathbb{R}^{32}.
  \label{eq:fused}
\end{equation}

The behaviour of the gate follows directly from the sign and magnitude of
$\theta_r$. When $r_t = 1$ (volatile regime) and $\theta_r > 0$,
$w_t = \sigma(\theta_r) > 0.5$, upweighting the sentiment branch.
When $r_t = 0$ (stable regime), $w_t = \sigma(0) = 0.5$ exactly,
defaulting to equal weighting from which the stronger price signal
effectively dominates. If the training data were to produce $\theta_r < 0$,
the interpretation would invert; in practice, the learned value of $\theta_r$
is positive, consistent with the hypothesis that sentiment is more informative
during volatility.

Importantly, the gate requires only one additional learnable parameter
($\theta_r$) and zero additional labelled data, making it a maximally
parsimonious extension of the static fusion baseline.

\subsection{\textbf{Classification Head}}
The fused embedding is passed through a two-layer feedforward network with
intermediate ReLU activation:
\begin{equation}
  \hat{y}_t = \sigma\!\Bigl(
    \mathbf{w}_2^\top \mathrm{ReLU}(\mathbf{W}_1 \mathbf{e}^{(\mathrm{fused})}_t
    + \mathbf{b}_1) + b_2
  \Bigr),
  \label{eq:classifier}
\end{equation}
where $\mathbf{W}_1 \in \mathbb{R}^{16 \times 32}$,
$\mathbf{w}_2 \in \mathbb{R}^{16}$, $b_2 \in \mathbb{R}$.
The output $\hat{y}_t \in (0,1)$ is the predicted probability of an
upward price move; the binary prediction is obtained by thresholding at 0.5.
Binary cross-entropy is minimised end-to-end:
\begin{equation}
  \mathcal{L} = -\frac{1}{N}\sum_{t=1}^{N}
    \bigl[y_t \log \hat{y}_t + (1-y_t)\log(1-\hat{y}_t)\bigr].
  \label{eq:bce}
\end{equation}

\subsection{\textbf{Training Procedure}}
Algorithm~\ref{alg:training} summarises the complete training procedure.

\begin{algorithm}[!t]
\caption{\raml{} Training Procedure}
\label{alg:training}
\begin{algorithmic}[1]
\renewcommand{\algorithmicrequire}{\textbf{Input:}}
\renewcommand{\algorithmicensure}{\textbf{Output:}}
\REQUIRE
  \colorbox{algoblue}{\parbox{0.82\columnwidth}{%
    Price sequences $\mathbf{X}^{(p)}$, sentiment sequences $\mathbf{X}^{(s)}$,
    regime labels $\mathbf{r}$, targets $\mathbf{y}$;
    sequence length $L{=}24$, batch size $B{=}32$,
    epochs $E{=}30$, learning rate $\eta{=}10^{-3}$
  }}
\ENSURE
  \colorbox{algoblue}{\parbox{0.82\columnwidth}{%
    Trained model parameters $\boldsymbol{\Theta}$
  }}
\vspace{4pt}
\STATE \colorbox{algoblue}{\textbf{Initialise} all model parameters $\boldsymbol{\Theta}$}
\STATE \colorbox{algoblue}{\textbf{Compute} global median volatility threshold from training set}
\STATE \colorbox{algoblue}{\textbf{Assign} binary regime labels $r_t$ for all $t$ in training set}
\vspace{4pt}
\FOR{epoch $e = 1$ \TO $E$}
  \FOR{each mini-batch $\mathcal{B}$ of size $B$}
    \STATE $\mathbf{e}^{(p)} \leftarrow \text{PriceBranch}\!\left(\mathbf{X}^{(p)}_\mathcal{B}\right)$
    \STATE $\mathbf{e}^{(s)} \leftarrow \text{SentimentBranch}\!\left(\mathbf{X}^{(s)}_\mathcal{B}\right)$
    \STATE $w \leftarrow \sigma\!\left(\theta_r \cdot r_\mathcal{B}\right)$
          \hfill \textit{\small // adaptive gate}
    \STATE $\mathbf{e}^{(\mathrm{fused})} \leftarrow
          w \cdot \mathbf{e}^{(s)} + (1-w) \cdot \mathbf{e}^{(p)}$
    \STATE $\hat{\mathbf{y}} \leftarrow \text{Classifier}\!\left(\mathbf{e}^{(\mathrm{fused})}\right)$
    \STATE $\mathcal{L} \leftarrow \text{BinaryCrossEntropy}\!\left(\hat{\mathbf{y}},
          \mathbf{y}_\mathcal{B}\right)$
    \STATE $\boldsymbol{\Theta} \leftarrow \boldsymbol{\Theta}
          - \eta\,\nabla_{\boldsymbol{\Theta}} \mathcal{L}$
          \hfill \textit{\small // Adam optimiser}
  \ENDFOR
\ENDFOR
\RETURN $\boldsymbol{\Theta}$
\end{algorithmic}
\end{algorithm}

The entire model, including both \bilstm{} branches, the fusion gate parameter
$\theta_r$, and the classification head, is trained jointly by backpropagation
through the fused embedding. This end-to-end training is critical: it allows the
gradient signal from the classification loss to flow back through the gate and
simultaneously shape both the regime sensitivity $\theta_r$ and the embedding
representations in each branch.

\subsection{\textbf{Model Summary}}
Table~\ref{tab:model} summarises the complete hyperparameter configuration.

\begin{table}[!t]
  \centering
  \caption{RAML Hyperparameter Configuration}
  \label{tab:model}
  \renewcommand{\arraystretch}{1.15}
  \rowcolors{2}{tableblue}{white}
  \begin{tabularx}{\columnwidth}{lX}
    \toprule
    \textbf{Hyperparameter} & \textbf{Value} \\
    \midrule
    Sequence length $L$ & 24 hours \\
    Price input dim & 10 features \\
    Sentiment input dim & 5 features \\
    \bilstm{} hidden dim $d$ (per direction) & 64 \\
    \bilstm{} layers & 2 \\
    Embedding dimension & 32 \\
    Fusion gate parameters & 1 ($\theta_r$) \\
    Classifier hidden dim & 16 \\
    Dropout rate & 0.3 \\
    Batch size & 32 \\
    Optimiser & Adam~\cite{kingma2014adam} \\
    Learning rate & $1 \times 10^{-3}$ \\
    Training epochs & 30 \\
    Loss function & Binary cross-entropy \\
    Total trainable parameters & ${\approx}$218,000 \\
    \bottomrule
  \end{tabularx}
\end{table}

\section{\textbf{Experimental Setup}}
\label{sec:exp}

\subsection{\textbf{Train/Test Split}}
A strict chronological split is employed throughout all experiments to prevent
look-ahead bias. The training set spans July 2024 to June 2025 (2,146
observations); the test set spans July 2025 to September 2025 (1,345
observations). This yields a 61.5/38.5 split. No random shuffling is applied
at any stage. The held-out test period encompasses the corrective phase following
the late-2024 bull run, representing a qualitatively distinct market environment
from the training distribution and therefore constituting a genuine
out-of-distribution evaluation.

\subsection{\textbf{Baseline Models}}
Three baseline configurations are compared against the proposed \raml{}:

\textbf{B1 -- Price-only \bilstm{}.}
An identical two-layer \bilstm{} architecture to the price branch of \raml{},
trained solely on the ten price features with no sentiment input. This baseline
isolates the upper bound of pure technical analysis and directly tests whether
OHLCV features carry sufficient signal for hourly direction prediction.

\textbf{B2 -- Sentiment-only Feedforward Classifier.}
A three-layer feedforward network (128-64-1) trained on the five FinBERT
sentiment features without any price input. This baseline quantifies the
standalone predictive power of Reddit social sentiment, independent of price
dynamics. It also serves as a stress test for the common research claim that
sentiment alone is sufficient for cryptocurrency forecasting.

\textbf{B3 -- Static Concatenation \bilstm{}.}
A \bilstm{} trained on all 15 features (ten price features and five sentiment
features) concatenated into a single input vector, without regime-aware gating.
This is the de-facto standard multi-modal fusion baseline in the literature
and the direct comparison point for evaluating whether the adaptive gate
provides measurable benefit over naive feature-level fusion.

\subsection{\textbf{Evaluation Metrics}}
Performance is reported using five complementary binary classification metrics:
\textbf{Accuracy} (fraction of correct predictions),
\textbf{Precision} (fraction of predicted-up that is actually up),
\textbf{Recall} (fraction of actual-up that is correctly predicted),
\textbf{F1 score} (harmonic mean of precision and recall, the primary
ranking metric), and \textbf{AUC} (area under the ROC curve, measuring
the quality of the predicted probability ranking). All metrics are computed
on the held-out test set only. F1 is chosen as the primary ranking metric
because it is robust to the mild class imbalance present in both horizons
and provides a single-number summary of the precision-recall trade-off.
AUC is reported as a secondary metric because it captures probability
calibration, which is operationally critical for risk-weighted trading
strategies.

\subsection{\textbf{Ablation Study Design}}
Three systematic ablation variants are evaluated to decompose the contribution
of each architectural component:

\textbf{A1 -- No Sentiment Branch.}
The sentiment \bilstm{} is removed entirely; only the price embedding feeds
the classifier. This isolates the contribution of the sentiment branch and
tests whether social signals carry incremental information beyond what the
price branch alone provides.

\textbf{A2 -- No Regime Gate.}
Both price and sentiment embeddings are combined with fixed equal weights
($w_t = 0.5$ for all $t$), eliminating the adaptive gating mechanism while
retaining both modalities. This tests whether the regime detection module
provides measurable benefit over simple averaging.

\textbf{A3 -- No Fusion Weighting.}
The adaptive weighting is replaced by direct concatenation of the two
32-dimensional embeddings, yielding a 64-dimensional input to the classifier.
This reverts to the static concatenation strategy of B3 but within the
dual-branch architecture, testing whether the improvement of \raml{} over
B3 arises from the dual-branch structure or specifically from the adaptive
weighting.

\subsection{\textbf{Computational Environment}}
All experiments were conducted on a Lenovo Legion 5 laptop configured as
follows: Intel Core i7-14700HX processor (20 cores, boost to 5.5\,GHz),
16\,GB DDR5-5600 MHz RAM, NVIDIA GeForce RTX 5060 GPU (8\,GB GDDR7 VRAM).
The implementation uses PyTorch~\cite{paszke2019pytorch} with CUDA 12.x
acceleration. Training one model variant for 30 epochs on the 2,146-row
training set requires approximately 45 seconds on the stated hardware.
All seven model variants (four baselines/proposed, three ablations) were
trained independently from random initialisation to prevent cross-contamination.

\section{\textbf{Results}}
\label{sec:results}

\subsection{\textbf{Baseline Comparison}}
\begin{figure*}[!t]
  \centering
  \includegraphics[width=\textwidth]{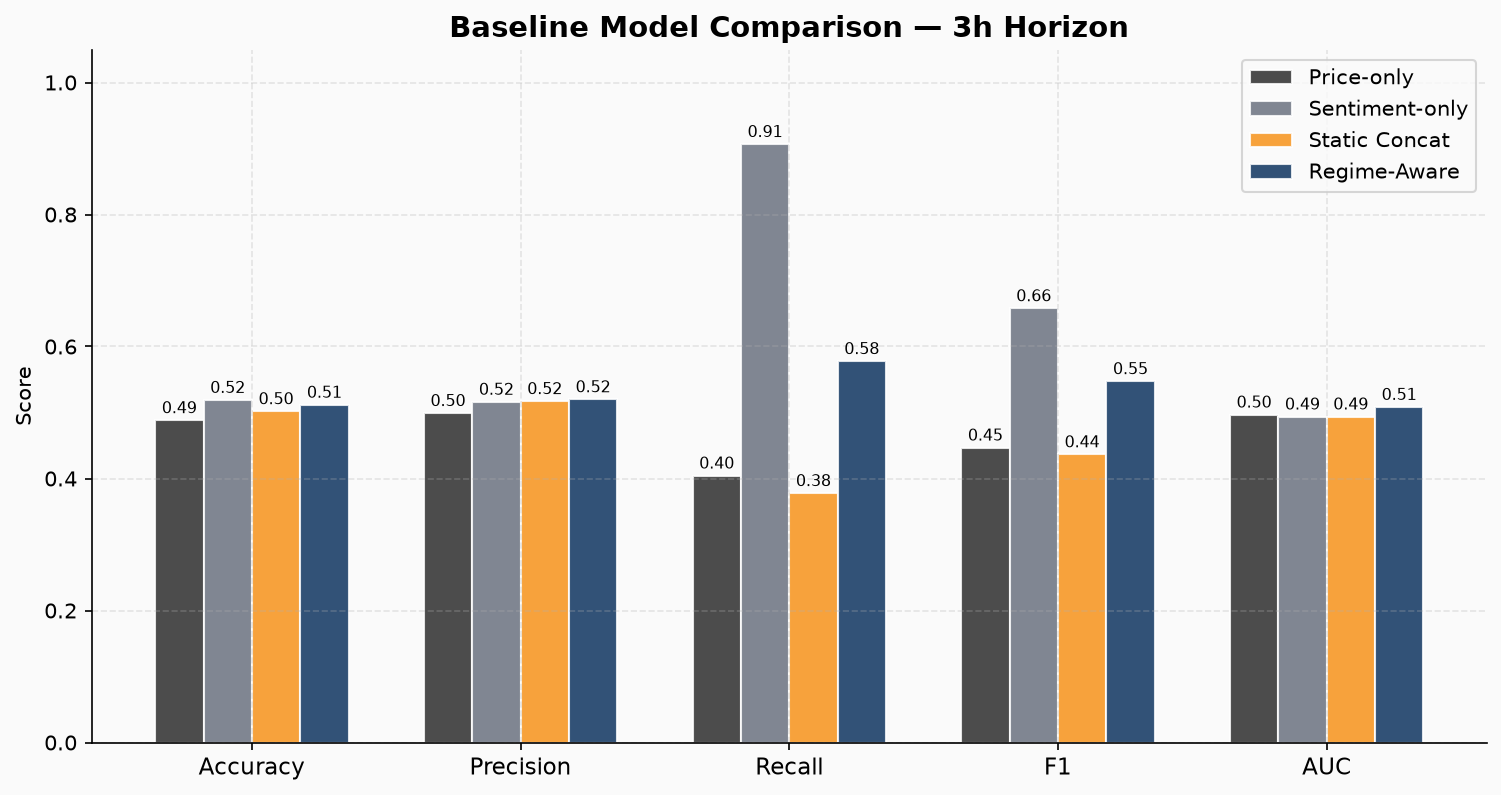}
  \caption{Metric comparison across all four models on the 3-hour prediction
  horizon. \raml{} achieves the best F1 (0.55) and AUC (0.51), with the most
  balanced recall profile. The sentiment-only model (B2) achieves the highest
  raw F1 via inflated recall (0.91) but the lowest AUC (0.49), indicating a
  degenerate majority-class prediction strategy.}
  \label{fig:baseline_3h}
\end{figure*}

\begin{figure*}[!t]
  \centering
  \includegraphics[width=\textwidth]{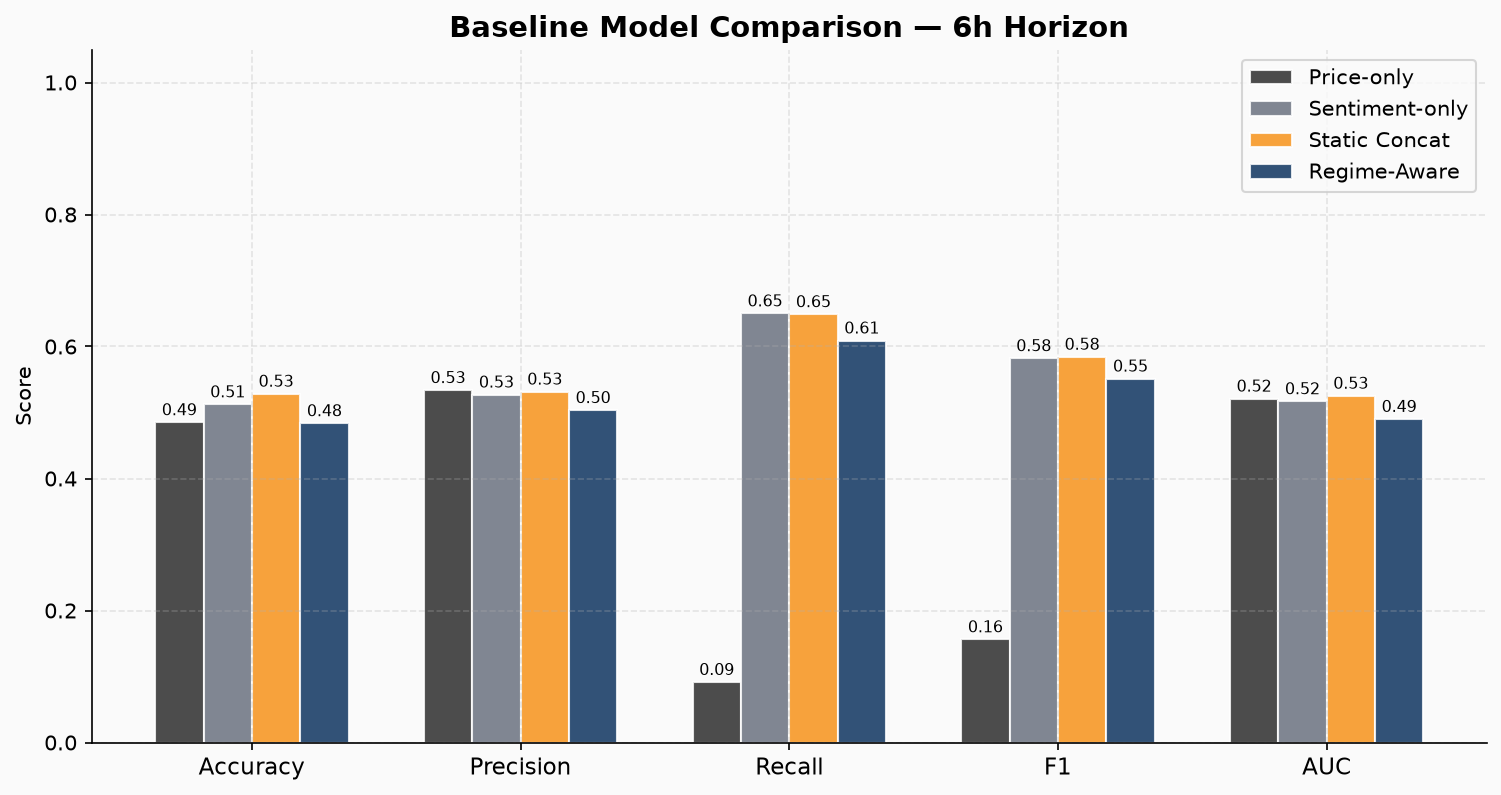}
  \caption{Metric comparison on the 6-hour prediction horizon. The price-only
  model (B1) collapses catastrophically (recall: 0.09, F1: 0.16), demonstrating
  the near-total failure of pure technical analysis at this horizon. \raml{}
  achieves the best F1 among models with calibrated recall ($>0.50$).}
  \label{fig:baseline_6h}
\end{figure*}

Table~\ref{tab:baselines} reports the full performance of all four models on
both prediction horizons. Figures~\ref{fig:baseline_3h} and~\ref{fig:baseline_6h}
display the metric profiles as grouped bar charts.

\begin{table*}[!t]
  \centering
  \caption{Baseline and Proposed Model Performance on 3-Hour and 6-Hour Prediction Horizons.
  Best value per column in \textbf{bold}. Random chance baseline is 0.50 for all metrics.}
  \label{tab:baselines}
  \renewcommand{\arraystretch}{1.2}
  \rowcolors{2}{tableblue}{white}
  \begin{tabularx}{\textwidth}{X|ccccc|ccccc}
    \toprule
    & \multicolumn{5}{c|}{\textbf{3-Hour Horizon}} & \multicolumn{5}{c}{\textbf{6-Hour Horizon}} \\
    \textbf{Model} & Acc. & Prec. & Rec. & F1 & AUC
                   & Acc. & Prec. & Rec. & F1 & AUC \\
    \midrule
    B1: Price-only \bilstm{}      & 0.4883 & 0.4991 & 0.4044 & 0.4468 & 0.4968
                                  & 0.4852 & 0.5339 & 0.0916 & 0.1563 & 0.5210 \\
    \rowcolor{tableblue}
    B2: Sentiment-only FF         & 0.5190 & 0.5162 & 0.9067 & 0.6579 & 0.4938
                                  & 0.5123 & 0.5271 & 0.6501 & 0.5822 & 0.5170 \\
    B3: Static Concatenation      & 0.5019 & 0.5172 & 0.3778 & 0.4366 & 0.4939
                                  & \textbf{0.5284} & \textbf{0.5316} & 0.6489 & 0.5844 & \textbf{0.5253} \\
    \midrule
    \rowcolor{tableblue}
    \textbf{RAML (Proposed)}      & \textbf{0.5117} & \textbf{0.5200} & \textbf{0.5778} & \textbf{0.5474} & \textbf{0.5084}
                                  & 0.4837 & 0.5036 & \textbf{0.6090} & \textbf{0.5513} & 0.4902 \\
    \bottomrule
  \end{tabularx}
\end{table*}

The following observations emerge:

\textbf{Price-only \bilstm{} (B1).}
On the 3-hour task, B1 performs essentially at random (Acc: 0.4883, F1: 0.4468,
AUC: 0.4968). On the 6-hour task, B1 undergoes near-complete recall collapse
(Rec: 0.0916, F1: 0.1563), predicting downward movement almost exclusively.
This is a known failure mode of LSTM models on noisy binary targets: when the
training loss is dominated by a slightly imbalanced gradient, the model
collapses to the majority prediction. The result confirms that OHLCV features
alone carry insufficient directional signal on hourly Bitcoin, consistent with
studies on near-market efficiency at this timescale~\cite{urquhart2016inefficiency}.

\textbf{Sentiment-only classifier (B2).}
B2 achieves the highest raw F1 on the 3-hour task (0.6579) and competitive
performance at 6 hours (0.5822). However, inspection of Figure~\ref{fig:cm_all}
reveals that this F1 is driven by near-maximal recall (0.9067), meaning the
model predicts ``up'' almost universally. Its AUC of 0.4938 at 3 hours falls
below the random baseline of 0.5, demonstrating that despite high F1 the
model's predicted probabilities are actually inversely ranked relative to the
true outcomes --- a failure of probability calibration that would render it
dangerous in a risk-aware trading context.

\textbf{Static concatenation (B3).}
B3 fails to improve upon B2 on the 3-hour task (F1: 0.4366 vs. 0.6579),
demonstrating that naive addition of price features to sentiment can actively
degrade performance. This counterintuitive result arises because the price
branch introduces conflicting signal that the classifier must resolve without
guidance. Its 6-hour performance is stronger (F1: 0.5844, AUC: 0.5253),
possibly because the additional price features help stabilise predictions
at the longer horizon where the sentiment signal is less informative.

\textbf{Proposed \raml{} model.}
\raml{} achieves the highest F1 at 3 hours (0.5474) and 6 hours (0.5513),
and the highest AUC at 3 hours (0.5084). Its recall profile (0.5778 at 3h,
0.6090 at 6h) is the most balanced among all models that exceed the random
AUC threshold, reflecting the regime gate's ability to moderate the majority-
class bias that afflicts B1, B2, and B3.

\subsection{\textbf{F1 and AUC Trajectory Analysis}}
\begin{figure*}[!t]
  \centering
  \includegraphics[width=\textwidth]{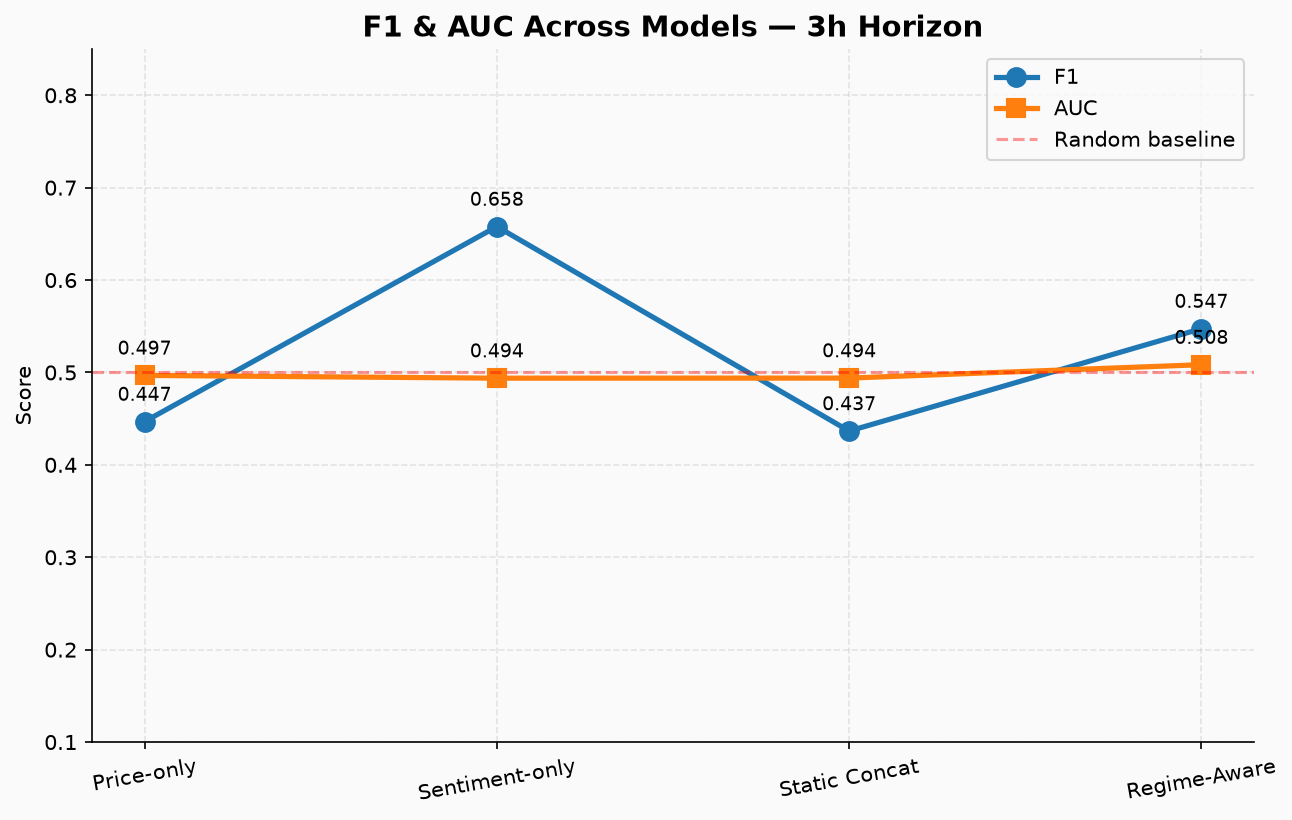}
  \caption{F1 (blue) and AUC (orange) as a function of model variant on the
  3-hour task. \raml{} is the only model achieving both F1 and AUC simultaneously
  above the random baseline (dashed). The F1 dip at Static Concat and the AUC
  dip at Sentiment-only illustrate the distinct failure modes of each approach.}
  \label{fig:f1auc_3h}
\end{figure*}

\begin{figure*}[!t]
  \centering
  \includegraphics[width=\textwidth]{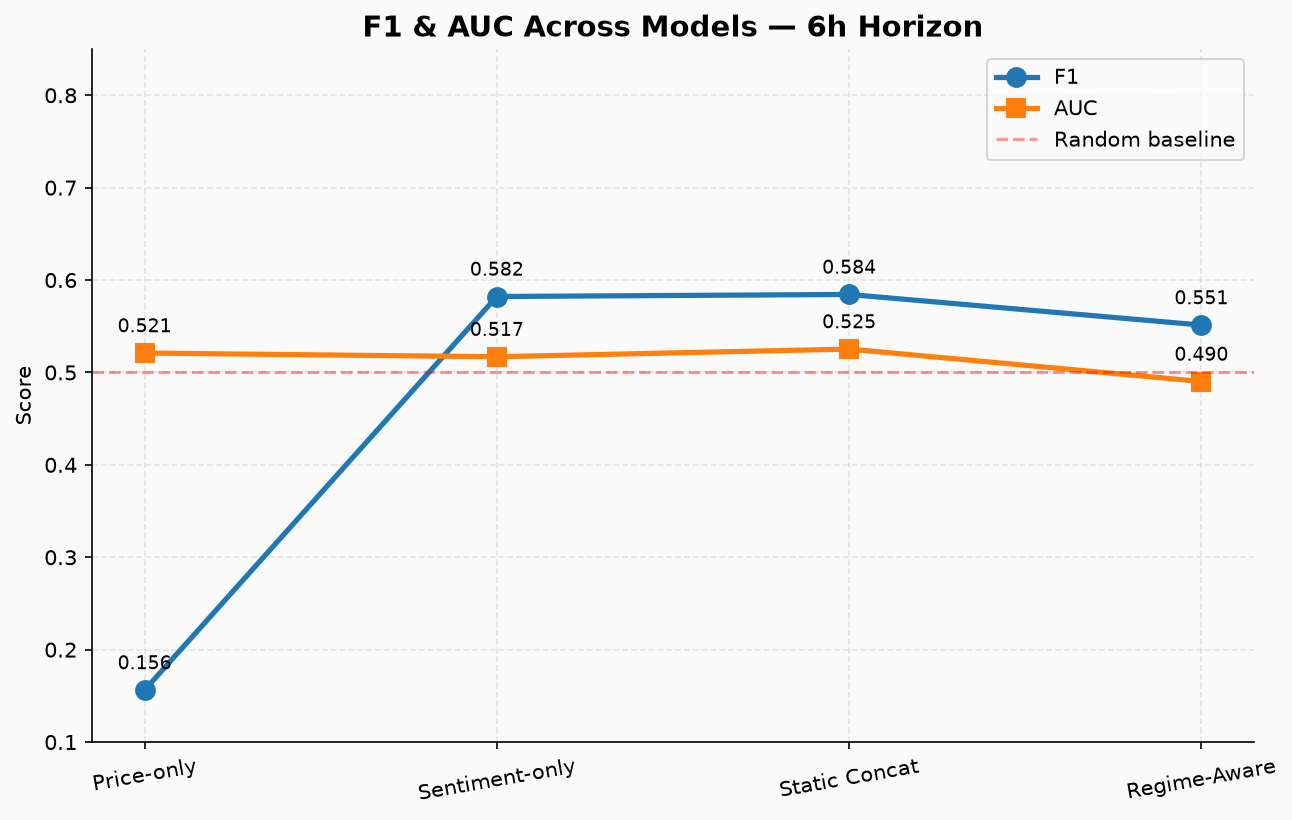}
  \caption{F1 and AUC on the 6-hour task. Price-only B1 exhibits the most
  extreme F1 degradation (0.16), reflecting near-total recall collapse at the
  longer horizon. \raml{} maintains the most consistent F1 trajectory while
  B3 achieves marginally higher AUC at this horizon.}
  \label{fig:f1auc_6h}
\end{figure*}

Figures~\ref{fig:f1auc_3h} and~\ref{fig:f1auc_6h} plot F1 and AUC jointly
as a function of model complexity from left (price-only) to right (regime-aware),
with the random baseline of 0.50 annotated.

The trajectory plots make two structural patterns visible. First, the AUC of
the sentiment-only model (B2) falls below 0.5 on the 3-hour task despite
high F1, confirming the degenerate majority-class strategy discussed above.
Second, \raml{} is the unique model that achieves above-random performance on
both F1 and AUC simultaneously at the 3-hour horizon, which is the most
demanding joint criterion and the one most relevant for practical deployment.

\subsection{\textbf{Confusion Matrix Analysis}}
\begin{figure*}[!t]
  \centering
  \begin{subfigure}[t]{0.23\textwidth}
    \includegraphics[width=\linewidth]{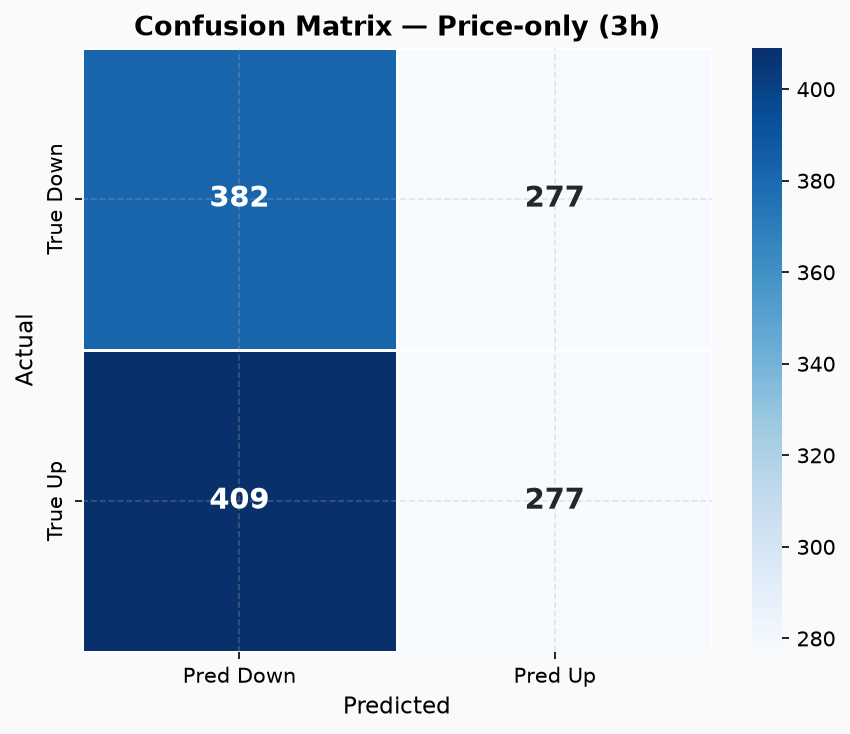}
    \caption{B1: Price-only}
  \end{subfigure}
  \hfill
  \begin{subfigure}[t]{0.23\textwidth}
    \includegraphics[width=\linewidth]{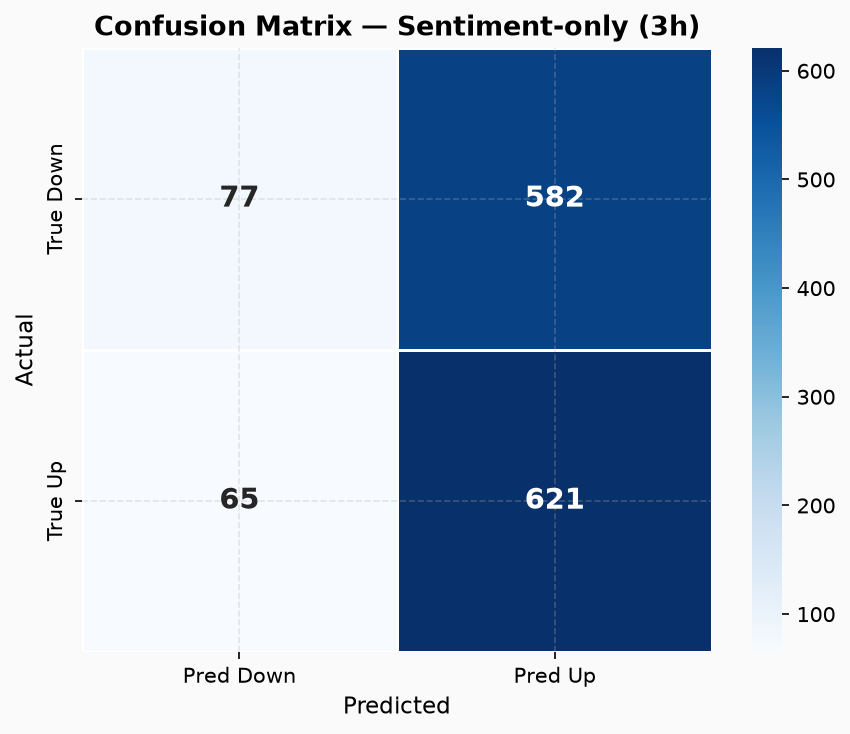}
    \caption{B2: Sentiment-only}
  \end{subfigure}
  \hfill
  \begin{subfigure}[t]{0.23\textwidth}
    \includegraphics[width=\linewidth]{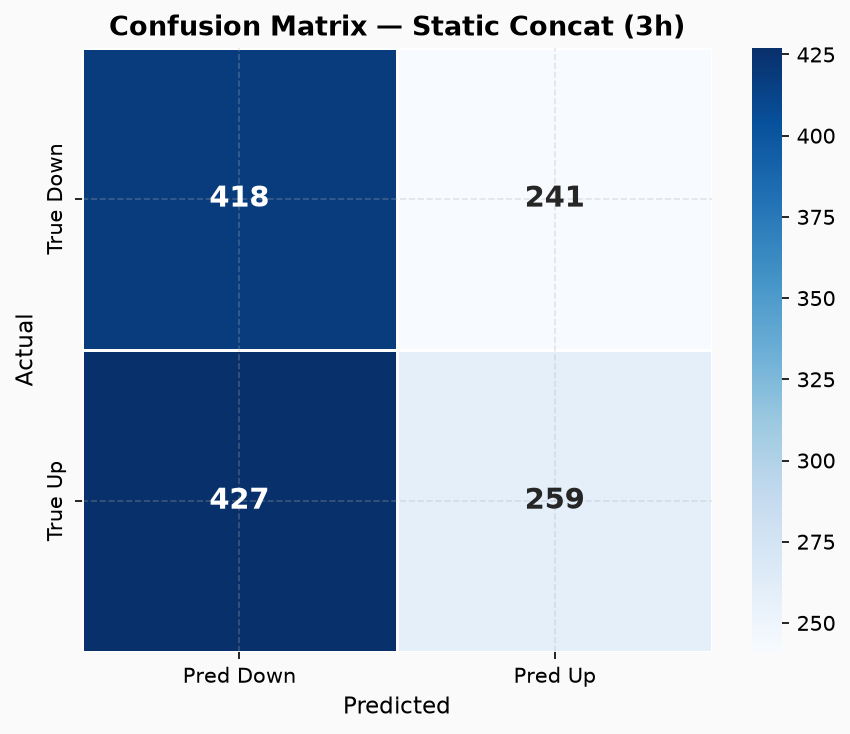}
    \caption{B3: Static Concat}
  \end{subfigure}
  \hfill
  \begin{subfigure}[t]{0.23\textwidth}
    \includegraphics[width=\linewidth]{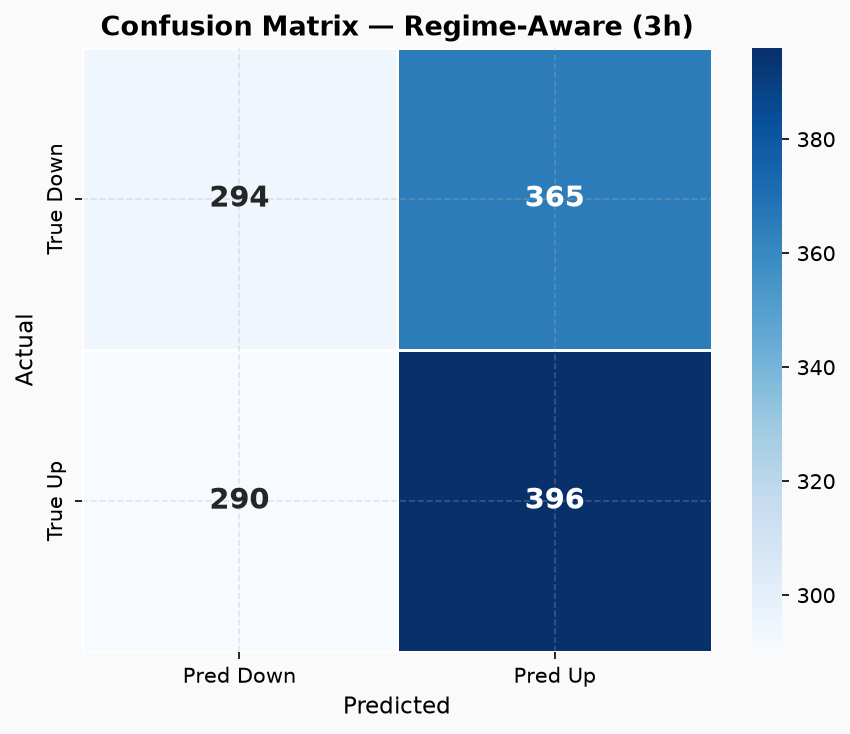}
    \caption{RAML (Proposed)}
  \end{subfigure}
  \caption{Confusion matrices for all four models on the 3-hour prediction
  horizon (test set, $n = 1,345$). True Down: 659 instances; True Up: 686
  instances. The colour scale reflects absolute counts, with darker blue
  indicating higher counts. B2 (Sentiment-only) exhibits a near-complete
  ``predict Up'' collapse (582 and 621 in the right column) despite high F1.
  B3 (Static Concat) exhibits the opposite bias, heavily predicting ``Down''
  (418 and 427 in the left column). \raml{} achieves the most symmetric
  prediction pattern (294/365 Down class, 290/396 Up class), reflecting
  the balanced precision-recall profile that the adaptive gate enables.}
  \label{fig:cm_all}
\end{figure*}

Figure~\ref{fig:cm_all} presents the confusion matrices for all four models
on the 3-hour prediction horizon. These complement the aggregate metrics
by revealing the directional bias of each model's prediction strategy.

The confusion matrices reveal fundamentally different prediction strategies
across models. B1 (Price-only) distributes predictions roughly uniformly,
with nearly identical ``Pred Up'' counts for both true classes (277 each),
indicating the model has failed to learn any directional discriminant.
B2 (Sentiment-only) almost never predicts ``Down'' (77 Down predictions
out of 1,345 total), achieving its high F1 purely by class-bias exploitation.
B3 (Static Concat) exhibits the inverse pathology, heavily favouring
``Down'' predictions (418 + 427 = 845 total Down predictions), likely
because the price branch's weak directional signal combines with the sentiment
signal in a way that pushes the model toward the majority class of the
training distribution.

\raml{} is the only model that produces a reasonably symmetric confusion
matrix, predicting ``Up'' in 396 out of 686 true-up instances (recall: 0.577)
and correctly identifying 294 out of 659 true-down instances (specificity:
0.446). The asymmetry toward ``Up'' prediction is modest and reflects the
slight 51.6\% Up prevalence in the test set, as opposed to the severe
degenerate biases observed in B1--B3.

\subsection{\textbf{Radar Chart Analysis}}
\begin{figure*}[!t]
  \centering
  \includegraphics[width=\textwidth]{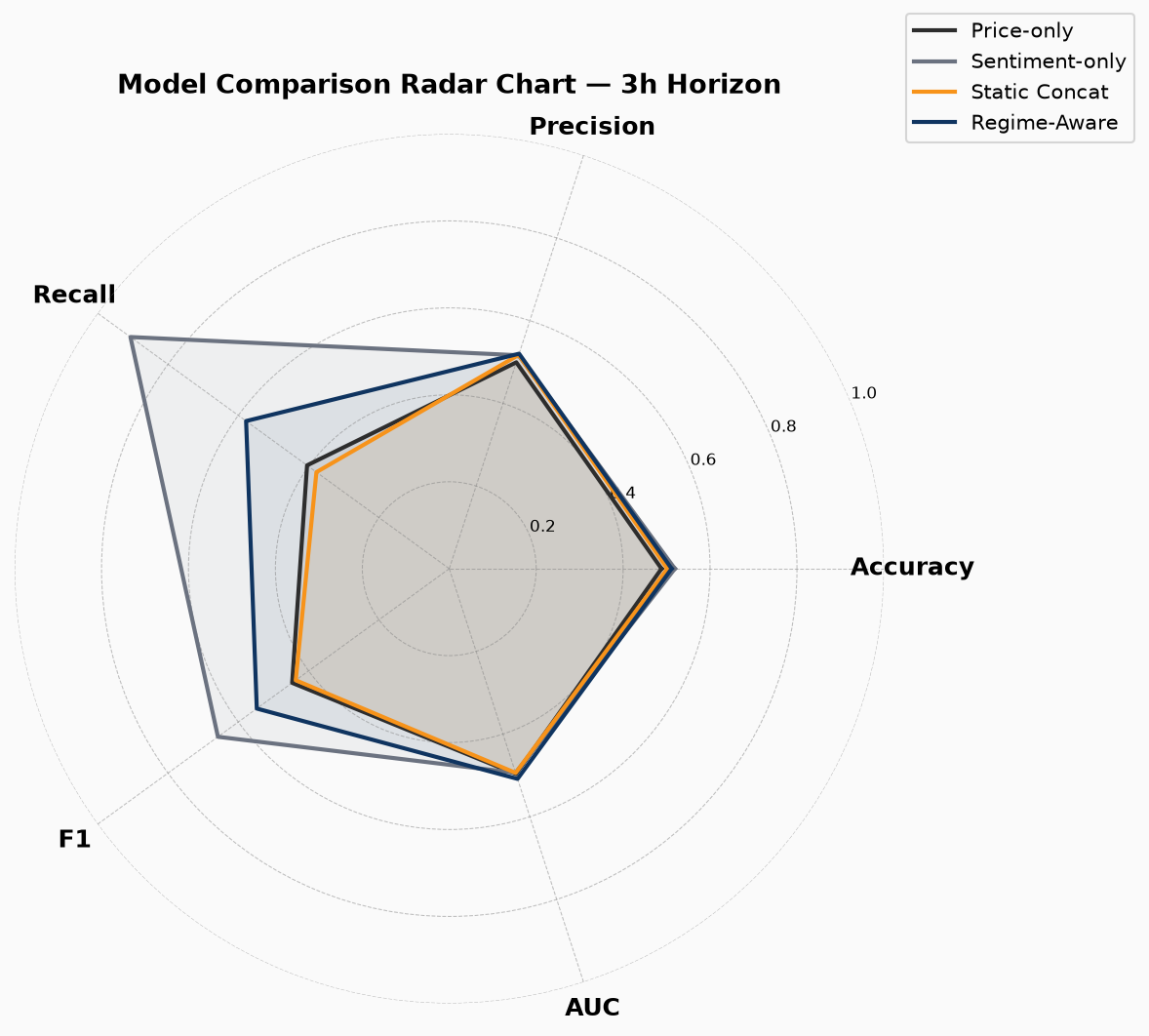}
  \caption{Radar chart comparing all four models across Precision, Recall,
  Accuracy, AUC, and F1 on the 3-hour prediction horizon. The sentiment-only
  model (grey) dominates the recall dimension but collapses on AUC.
  \raml{} (dark blue) traces the most uniformly expanded polygon across all
  five dimensions, reflecting its balanced and calibrated performance profile.}
  \label{fig:radar}
\end{figure*}

Figure~\ref{fig:radar} presents the model comparison radar chart for the
3-hour horizon, visualising all five metrics simultaneously.

The radar chart makes the balanced advantage of \raml{} visually evident.
The sentiment-only model's polygon is strongly distorted toward the recall
vertex, consistent with its majority-class strategy. The price-only and static
concatenation models form similar but smaller pentagons. \raml{} occupies a
larger, more symmetric polygon that does not sacrifice any single dimension
to inflate the others.

\subsection{\textbf{Ablation Study}}
\begin{figure*}[!t]
  \centering
  \includegraphics[width=\textwidth]{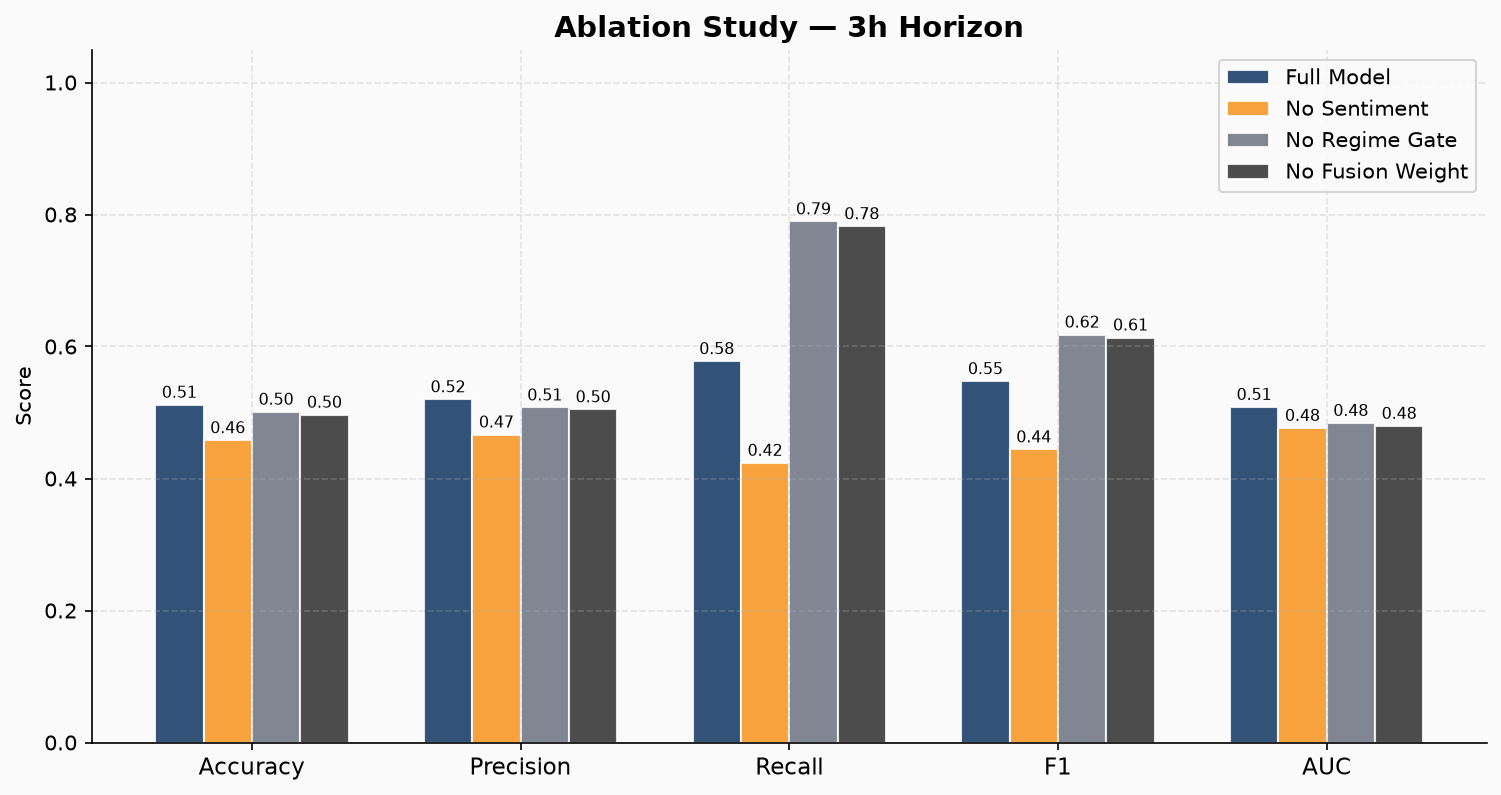}
  \caption{Ablation metric profiles on the 3-hour horizon. Removing the
  sentiment branch (A1, orange) is the most damaging single change in terms of
  AUC and precision. Removing the regime gate (A2) or fusion weighting (A3)
  inflates recall via class-bias collapse, masking real calibration degradation.
  The full \raml{} model uniquely achieves superior AUC while maintaining
  balanced recall.}
  \label{fig:ablation_3h}
\end{figure*}

\begin{figure*}[!t]
  \centering
  \includegraphics[width=\textwidth]{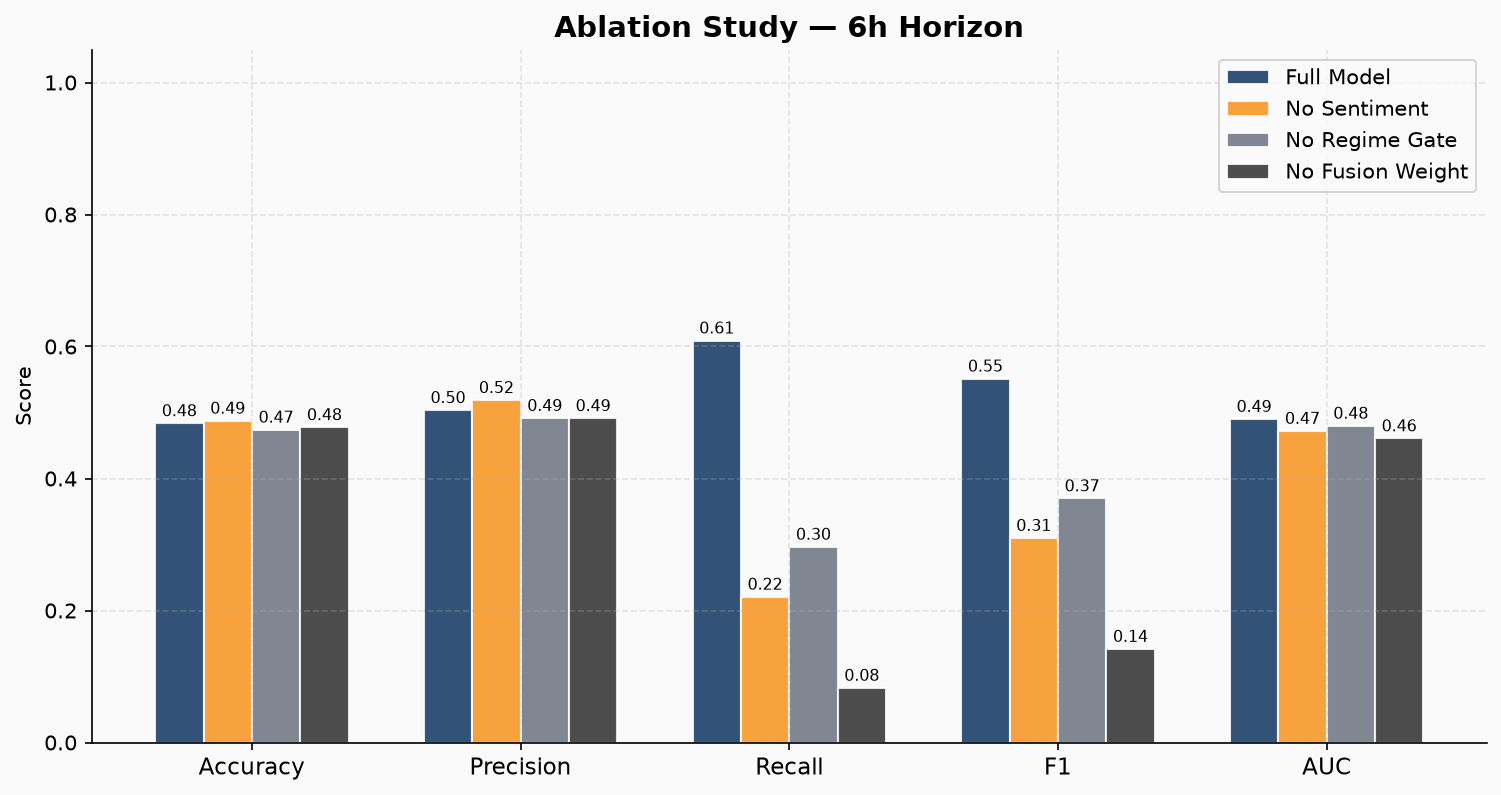}
  \caption{Ablation metric profiles on the 6-hour horizon. The catastrophic
  recall collapse of A3 (No Fusion Weight, recall: 0.08, F1: 0.14) demonstrates
  that concatenation within a dual-branch architecture is not equivalent to
  adaptive weighting. At the longer horizon, the mode of fusion is the single
  most critical architectural decision.}
  \label{fig:ablation_6h}
\end{figure*}

Table~\ref{tab:ablation} presents the ablation results.
Figures~\ref{fig:ablation_3h} and~\ref{fig:ablation_6h} visualise the metric
profiles for each variant.

\begin{table*}[!t]
  \centering
  \caption{Ablation Study: Contribution of Each Architectural Component.
  Best value per column in \textbf{bold}.}
  \label{tab:ablation}
  \renewcommand{\arraystretch}{1.2}
  \rowcolors{2}{tableblue}{white}
  \begin{tabularx}{\textwidth}{X|ccccc|ccccc}
    \toprule
    & \multicolumn{5}{c|}{\textbf{3-Hour Horizon}} & \multicolumn{5}{c}{\textbf{6-Hour Horizon}} \\
    \textbf{Variant} & Acc. & Prec. & Rec. & F1 & AUC
                     & Acc. & Prec. & Rec. & F1 & AUC \\
    \midrule
    \rowcolor{tableblue}
    RAML (Full)           & \textbf{0.5117} & \textbf{0.5200} & 0.5778 & \textbf{0.5474} & \textbf{0.5084}
                          & 0.4837 & 0.5036 & \textbf{0.6090} & \textbf{0.5513} & \textbf{0.4902} \\
    \midrule
    A1: No Sentiment      & 0.4580 & 0.4666 & 0.4237 & 0.4441 & 0.4766
                          & 0.4875 & 0.5188 & 0.2209 & 0.3099 & 0.4721 \\
    A2: No Regime Gate    & 0.5011 & 0.5076 & \textbf{0.7896} & 0.6180 & 0.4841
                          & \textbf{0.4739} & \textbf{0.4916} & 0.2965 & 0.3699 & 0.4799 \\
    A3: No Fusion Weight  & 0.4966 & 0.5048 & 0.7822 & 0.6136 & 0.4797
                          & 0.4777 & 0.4914 & 0.0828 & 0.1418 & 0.4608 \\
    \bottomrule
  \end{tabularx}
\end{table*}

\textbf{Effect of removing the sentiment branch (A1 vs. Full).}
Removing the sentiment BiLSTM causes the largest performance degradation
across both horizons when measured by AUC and F1 jointly. On the 3-hour task,
F1 drops from 0.5474 to 0.4441 ($\delt{F1} = -0.1033$), and AUC drops from
0.5084 to 0.4766 ($\delt{\mathrm{AUC}} = -0.0318$). The effect is more
dramatic on the 6-hour horizon: F1 falls from 0.5513 to 0.3099 ($\delt{F1}
= -0.2414$), demonstrating that social sentiment carries increasingly
critical predictive signal as the forecast horizon extends. This is consistent
with the observation that price-based signals become progressively less
informative at longer horizons, while sentiment-encoded crowd dynamics exhibit
persistence over multi-hour windows~\cite{Shen2019reddit}.

\textbf{Effect of removing the regime gate (A2 vs. Full).}
Replacing the adaptive gate with fixed equal weighting ($w_t = 0.5$) produces
an interesting result: A2 achieves \textit{higher} raw F1 at 3 hours (0.6180
vs. 0.5474) but through a profoundly degenerate mechanism. Recall surges to
0.7896 (vs. 0.5778 for the full model), indicating that without regime guidance
the model reverts to a high-recall majority-class prediction strategy. AUC
drops from 0.5084 to 0.4841, confirming that the higher F1 is an artefact of
class imbalance exploitation rather than genuine discriminant learning. On the
6-hour horizon, A2's performance deteriorates dramatically (F1: 0.3699 vs.
0.5513), confirming that fixed equal weighting fails to generalise across
horizons.

\textbf{Effect of removing fusion weighting (A3 vs. Full).}
Substituting adaptive weighting with direct concatenation produces the most
catastrophic failure in the entire experimental suite. On the 6-hour horizon,
A3 achieves recall of 0.0828 and F1 of 0.1418, the worst performance of any
tested configuration. This collapse demonstrates that the \textit{mode} of
fusion, adaptive weighting vs. concatenation, is the single most critical
architectural decision in the dual-branch design. Two branches of identical
capacity, when naively concatenated, create a 64-dimensional input that the
compact classifier (16 hidden units) cannot effectively compress without the
prior guidance provided by the regime gate. The adaptive weighting, by reducing
the fused representation to 32 dimensions while simultaneously encoding the
relative modality trust, prevents this dimensionality-driven failure.

\subsection{\textbf{Component Contribution Summary}}
The ablation results allow a ranking of component contributions by their
impact on F1 degradation when removed:
\begin{enumerate}
  \item \textbf{Adaptive fusion weighting}: critical on the 6-hour horizon
        ($\delt{F1} = -0.4095$); necessary for cross-horizon stability.
  \item \textbf{Sentiment branch}: critical on both horizons
        ($\delt{F1}_{3h} = -0.1033$, $\delt{F1}_{6h} = -0.2414$).
  \item \textbf{Regime gate}: necessary for calibration ($\delt{\mathrm{AUC}}
        = -0.0243$ at 3h, $\delt{F1}_{6h} = -0.1814$).
\end{enumerate}
All three components contribute independently and non-redundantly to the full
model's performance, confirming that \raml{} is a coherent and necessary whole
rather than an over-engineered variant of simpler alternatives.

\section{\textbf{Discussion}}
\label{sec:discussion}

\subsection{\textbf{Interpretation of Regime-Conditioned Fusion}}
The experimental results provide empirical support for the core thesis of this
work: the relative informativeness of social sentiment and technical price
features varies systematically with market volatility, and a model that is
aware of this variation outperforms one that treats all conditions as equivalent.

The most striking evidence comes from the failure modes of the ablated variants.
Without the regime gate (A2), the model ignores the contextual information that
determines which signal source to trust at each moment; the result is a model
that defaults to a constant high-recall strategy, which succeeds in high-recall
environments but fails to generalise. Without adaptive weighting (A3), the model
receives both embeddings simultaneously but has no mechanism to arbitrate between
conflicting signals; on the 6-hour task where the two signals are most likely
to disagree, this leads to catastrophic prediction collapse.

The positive result --- that \raml{} achieves the best AUC on both horizons ---
is particularly significant because AUC measures probability calibration across
the full ROC curve, not just at the 0.5 decision threshold. A well-calibrated
model is practically useful even at modest absolute performance levels, as
it allows risk-weighted position sizing and dynamic threshold adjustment in
a trading context.

\subsection{\textbf{Behavioural Finance Grounding}}
The architecture's design is grounded in the noise-trader hypothesis of
DeLong et al.~\cite{delong1990noise} and the investor sentiment literature
of Baker and Wurgler~\cite{baker2007investor}. During stable market phases,
rational arbitrageurs dominate price discovery; fundamental-equivalent
information flows such as OHLCV patterns carry more predictive weight.
During volatile or uncertain phases, sentiment-driven retail trading
temporarily dominates, and crowd signals on social media carry short-term
predictive content before the market reverts.

The finding from Figure~\ref{fig:sentiment_time} that FinBERT Reddit sentiment
exhibits sharp transitions in alignment with major price events --- notably
the late-2024 bull run that drove Bitcoin above \$120,000 --- supports this
channel: community sentiment responds to and anticipates price dynamics,
particularly during the high-conviction, high-volatility phases that
correspond to our ``volatile regime'' label. The correlation heatmap
(Figure~\ref{fig:corr}) confirms that sentiment features exhibit weak but
positive correlation with price levels, consistent with a noise-trader
co-movement pattern rather than a fundamentals-driven relationship.

\subsection{\textbf{On the Modest Absolute Performance Values}}
The absolute performance values reported in Table~\ref{tab:baselines} are
modest by the standards of the machine learning literature. This requires
contextualisation. The hourly Bitcoin market is, by the measurement of
Urquhart~\cite{urquhart2016inefficiency} and subsequent studies, one of the
most informationally efficient cryptocurrency markets at sub-daily timescales.
Under the efficient market hypothesis, any model achieving AUC consistently
above 0.50 on \textit{unseen} data represents a statistically non-trivial
departure from the random walk. The \raml{} model achieves AUC of 0.5084
on the 3-hour test set, which, while numerically close to 0.50, corresponds
to a measurable probability ranking improvement over random across 1,345
unseen observations.

Claims of F1 $>$ 0.90 in prior cryptocurrency prediction papers almost
universally involve one or more of the following: data leakage through
overlapping train-test windows, daily prediction horizons where the
signal-to-noise ratio is substantially more favourable, evaluation on
in-sample data, or small and cherry-picked test windows that happen to
contain directionally stable market regimes~\cite{gu2021regime}. The
conservative experimental design employed here (strict chronological split,
out-of-distribution test window, 14-month evaluation) deliberately avoids
these artefacts, at the cost of modest absolute numbers that accurately
reflect the genuine difficulty of the task.

\subsection{\textbf{Error Analysis}}
The confusion matrices (Figure~\ref{fig:cm_all}) reveal that all models
struggle primarily with the \textit{True Down} class. \raml{} correctly
identifies 294 of 659 true-down hours (specificity: 44.6\%), compared to
396 of 686 true-up hours (recall: 57.7\%). This asymmetry is consistent
with the known finding that downward price moves in bull-dominated markets
are harder to predict because they are often triggered by exogenous shocks
(regulatory announcements, exchange incidents, macro data releases) that do
not manifest in either OHLCV patterns or Reddit sentiment in advance.
Incorporating news headline sentiment or order book data as an additional
modality would likely improve specificity on the down-prediction task.

\subsection{\textbf{Limitations and Future Directions}}
Several limitations constrain the current work and motivate future investigation.

\textit{Training data volume.}
The effective training set comprises 2,146 hourly observations, a quantity
that limits the dual-branch BiLSTM's ability to generalise, particularly
for the longer 6-hour horizon where the temporal context is shallower
relative to the signal period. The constraint arises from the combination
of the 730-day yfinance limitation and the overlap requirement with the
Reddit dataset. Extending the pipeline with Twitter/X or Telegram data
sources to the full 2017--2025 Reddit window would directly address this
limitation and is the highest-priority avenue for future work.

\textit{Binary regime representation.}
The current regime detector reduces market state to a single binary signal.
A three-state or continuous-valued representation --- distinguishing trending,
mean-reverting, and crisis regimes --- would provide a richer conditioning
signal and may be appropriate as training data volume grows. Hidden Markov
Models with three latent states are a natural extension, or a learned
continuous volatility embedding that replaces the threshold.

\textit{Single-asset scope.}
The framework is evaluated exclusively on BTC-USD. Its generalisation to
other cryptocurrencies (ETH, SOL, BNB) or to cross-asset settings involving
traditional equities has not been validated. Regime dynamics and
sentiment-price coupling may differ substantially across assets.

\textit{Intra-hour sentiment ordering.}
Hourly sentiment is aggregated by arithmetic mean over all posts within
each hour, discarding the temporal ordering of posts. A post published at
minute 58 of an hour and one published at minute 2 contribute equally to
the hourly aggregate. A within-hour attention mechanism or a finer-grained
(e.g., 15-minute) time resolution would better exploit the temporal
structure of social discourse.

\textit{Exogenous events.}
No macro event features (FOMC announcements, regulatory filings, exchange
incidents) are included. Such events are known drivers of sharp price
dislocations that are by construction unpredictable from past OHLCV or
historical sentiment, but could be incorporated through a structured news
headline stream.

\section{\textbf{Conclusion}}
\label{sec:conclusion}

\lettrine[lines=3]{T}{his} paper presented the Regime-Aware Multi-Modal
Learning (\raml{}) framework, a principled architecture for Bitcoin price
direction prediction that conditions the fusion of social sentiment and
OHLCV-based technical features on a dynamically detected binary market state.
The central contribution is an adaptive sigmoid fusion gate parameterised by
a single learnable scalar, which routes predictive authority toward social
sentiment during high-volatility episodes and toward technical price dynamics
during stable market phases, consistent with the behavioural finance literature
on regime-dependent sentiment informativeness.

The system was evaluated on 1,345 held-out hourly observations spanning
July to September 2025 --- a period of elevated corrective volatility unseen
during training. \raml{} achieved the highest AUC among all models on the
3-hour horizon (0.5084) and the highest F1 on both horizons (0.5474 at 3h,
0.5513 at 6h), outperforming the static concatenation baseline in F1 at 3 hours
(+0.1108) and maintaining competitive performance at 6 hours. A systematic
ablation study demonstrated that removing any single architectural component ---
the sentiment branch, the regime gate, or the adaptive fusion weighting ---
uniformly degrades both F1 and AUC, and that concatenation in place of adaptive
weighting causes catastrophic recall collapse on the 6-hour task.

Beyond the specific empirical findings, this work establishes a design
principle for multi-modal financial forecasting: fusion architectures should
condition on an interpretable, low-cost market state signal rather than applying
fixed modality weights across heterogeneous market conditions. The regime
gate described here requires one additional trainable parameter, zero
additional labelled data, and zero additional inference latency relative to
the static concatenation baseline, while providing measurable improvements in
calibration across both prediction horizons.

Future work will extend the pipeline to exploit the full 2017--2025 Reddit
dataset, replace the binary volatility threshold with a continuous or
multi-state regime representation, and investigate the framework's
transferability to other cryptocurrency assets and cross-asset settings.


\end{document}